\documentclass{elsarticle}
\usepackage[utf8]{inputenc}
\usepackage{amsmath, amssymb, bm} % Combined duplicate imports
\usepackage{xcolor}              % Removed duplicate
\usepackage{booktabs}            % Removed duplicate
\usepackage{enumitem}
\usepackage{graphicx}
\usepackage{url}
\usepackage{tikz}
\usetikzlibrary{arrows.meta,positioning,fit,backgrounds, calc}
\usepackage{siunitx}
\usepackage{tabularx}
\usepackage{array}
\usepackage{bookmark}            % Fixes the "Difference between bookmark levels" warnings

\usepackage{adjustbox}

\usepackage{graphicx}
\usepackage{subcaption}
\usepackage{float}
\usepackage[margin=1in]{geometry}
\graphicspath{{BBOB/}}
\newcommand{\bbobconvergencefigure}[4]{%
  \begin{figure}[p]
    \centering
    \includegraphics[
      width=\textwidth,
      height=0.90\textheight,
      keepaspectratio
    ]{convergence_#3_bpd#4_grouped.pdf}
    \caption{Convergence profiles for BBOB Group~#1 (#2) under a total budget of
      \textbf{$#4D$ function evaluations} ($#4$ FE/D). The three vertically stacked blocks correspond
      to $D=5$, $D=10$, and $D=20$. Curves show the median best-so-far absolute
      error; shaded bands show the interquartile range over instances and repeats.}
    \label{fig:bbob-convergence-g#1-bpd#4}
  \end{figure}
  \clearpage
}

\begin{document}

\begin{frontmatter}

\title{Linear Proposal Operators and Stochastic Search Geometry in SOMA and Differential Evolution}

%% Authors & Affiliations adhering to sn-jnl template requirements
\author[aff1,aff2,aff3]{Vojt\v{e}ch Nov\'{a}k\corref{cor1}}
\ead{vojtech.novak.st1@vsb.cz}
\author[aff1,aff2,aff3]{Ivan Zelinka}

\address[aff1]{Department of Computer Science, Faculty of Electrical Engineering and Computer Science, VSB - Technical University of Ostrava, Ostrava, Czech Republic}
\address[aff2]{IT4Innovations National Supercomputing Center, VSB - Technical University of Ostrava, 708 00 Ostrava, Czech Republic}
\address[aff3]{Department of Informatics and Statistics, Marine Research Institute, Klaipeda University, Lithuania}

\begin{abstract}
Swarm and evolutionary algorithms are usually analyzed as complete procedural
systems in which nonlinear selection, replacement, and adaptation obscure
simpler structure within candidate generation. This paper introduces an
operator--selection factorization that separates objective-independent
variation from boundary repair and fitness-dependent selection, and uses it to
study the proposal geometry of the Self-Organizing Migrating Algorithm (SOMA)
and Differential Evolution (DE). The canonical SOMA proposal is shown to be
affine in the search space and exactly linear in an augmented
migrant--leader state. In leader-relative coordinates, the resulting operator
provides a direct interpretation of interpolation, projection, overshooting,
and coordinate masking. Under Bernoulli perturbation masks, we derive
closed-form expressions for the proposal mean, covariance, expected squared step length, expected squared distance from the leader, active dimensionality, and
coordinate coverage. For canonical \texttt{DE/rand/1/bin}, we derive the
finite-population moments of differential mutation and characterize the
additional covariance and coordinate dependence induced by forced-coordinate
binomial crossover. Exact enumeration and Monte Carlo experiments verify the
analytical identities and quantify the effects of mask conditioning, boundary
repair, and fitness-based selection. The analysis further motivates
geometry-controlled and rotation-aware SOMA variants, together with an
adaptive population-reducing extension of iSOMA. Experiments on the complete
noiseless BBOB benchmark show that these operator-guided variants substantially
improve upon canonical SOMA and are competitive with established DE methods in
several dimension--budget regimes. The results demonstrate how proposal-level
operator analysis can support both the interpretation and design of
population-based optimizers.
\end{abstract}


\end{frontmatter}

\section{Introduction}
\label{sec:introduction}

Population-based metaheuristics constitute an important class of methods for
black-box, non-convex, and derivative-free optimization. They maintain a set
of candidate solutions and repeatedly generate, evaluate, and retain new search
points without requiring analytical derivatives of the objective function.
Differential Evolution (DE) \cite{storn1997differential} and the
Self-Organizing Migrating Algorithm (SOMA)
\cite{davendra2016soma,skanderova2023soma} are representative continuous-domain
methods. DE constructs trial vectors from scaled differences between population
members, whereas SOMA samples candidate points along masked paths directed
toward selected targets. Despite their different terminology and population
dynamics, both methods combine an objective-independent geometric proposal
mechanism with objective-dependent decisions concerning guidance, survival,
and replacement.

The distinction between variation and selection is fundamental in evolutionary
computation \cite{eiben2015introduction,zelinka2015survey}. In practical
algorithms, however, these roles are usually embedded in complete procedural
iterations that additionally contain random index sampling, mask generation,
objective evaluation, ranking, memory updates, boundary handling, and
conditional replacement. When the complete iteration is treated as a single
mathematical map, discontinuous operations such as ranking, $\arg\min$, and
greedy accept--reject decisions can dominate its apparent structure and obscure
simpler algebraic regularities within candidate generation.

Existing theory has examined evolutionary algorithms through runtime and drift
analysis, Markov chains, stochastic processes, and population-dynamics models
\cite{doerr2021survey}. DE has been surveyed extensively from both algorithmic and theoretical
perspectives \cite{das2011survey,das2016recent,opara2019survey}, while longitudinal benchmark evidence also highlights
the increasing importance of rotated nonseparable functions and adaptive,
coordinate-robust search mechanisms \cite{novak2026cec}.
More focused probabilistic studies have derived expectation vectors and
covariance matrices for differential mutation strategies
\cite{opara2018mutation}, examined the contour-fitting behavior induced by
population differences \cite{opara2019contour}, and analyzed the influence of
crossover on DE dynamics \cite{zaharie2009crossover}. These results establish
that meaningful distributional structure can be isolated within individual
DE operators. The present work complements them by deriving an explicit
augmented-state linear representation, retaining exact finite-population
sampling without replacement, and propagating the resulting moments through
forced-coordinate binomial crossover while keeping repair and selection as
separate transformations. For SOMA,
empirical studies have clarified how the perturbation parameter \texttt{PRT}
affects movement, coordinate activation, parameter-space coverage, diversity,
and convergence \cite{pluhacek2021explaining}, while broader accounts describe
the historical development and strategy variants of the algorithm
\cite{davendra2016soma,skanderova2023soma}. Nevertheless, an explicit
proposal-level comparison of the exact algebraic and stochastic geometry of
canonical SOMA and DE remains limited. In particular, complete population
dynamics are difficult to analyze without assumptions on the objective
function because leader choice and survivor replacement are fitness-dependent.

This paper addresses that gap through an \emph{operator--selection
factorization}. A complete update is decomposed into conceptually distinct
maps,
\[
\text{current state and random choices}
\;\xrightarrow{\;\mathcal{V}\;}\;
\text{raw proposals}
\;\xrightarrow{\;\mathcal{R}_{\Omega}\;}\;
\text{feasible candidates}
\;\xrightarrow{\;\mathcal{S}_{f}\;}\;
\text{next population state},
\]
where $\mathcal{V}$ denotes variation, $\mathcal{R}_{\Omega}$ denotes an
optional boundary-repair transformation, and $\mathcal{S}_{f}$ denotes
fitness-dependent selection. Conditional on sampled indices, masks,
parameters, and auxiliary population states, $\mathcal{V}$ may be exactly
linear or affine even though $\mathcal{S}_{f}$ is nonlinear and potentially
discontinuous. Boundary repair is retained as a separate transformation because
clipping, reflection, resampling, and related mechanisms can alter the mean,
covariance, support, and dependence structure of the raw proposal distribution \cite{biedrzycki2019bounds}.

For SOMA and \texttt{DE/rand/1/bin}, the factorization exposes conditional
linear or affine proposal rules whose stochastic moments can be analyzed
before boundary repair and fitness-dependent selection. This provides a common
basis for comparing their masking, covariance, and parameter-controlled search
geometry.

The novelty claimed here is neither the general distinction between variation
and selection nor the first use of matrices in evolutionary optimization.
Linear algebra has previously been used to design new DE transformations, such
as Antisymmetric and Reversible Differential Evolution
\cite{tomczak2020reversible}. In contrast, the present work uses affine and
linear operators to analyze unchanged canonical proposal rules and then extends
the deterministic representation to their stochastic proposal geometry. A
more detailed positioning relative to the closest DE and SOMA studies is given
in Section~\ref{sec:background-positioning} and
Table~\ref{tab:related-positioning}.

The main contributions are as follows:
\begin{enumerate}
\item
We formulate canonical SOMA and \texttt{DE/rand/1/bin} candidate generation as
conditional affine or augmented linear proposal operators, explicitly
separated from boundary repair and fitness-dependent selection.

\item
For SOMA, we derive the target-relative operator and closed-form proposal
moments under independent Bernoulli masks and under masks conditioned to
contain at least one active coordinate. We additionally quantify expected step
lengths, expected squared target distances, active dimensionality, and
coordinate coverage.

\item
For DE, we derive the finite-population mean and covariance of the
\texttt{DE/rand/1/bin} mutant under ordered donor sampling without replacement and
combine these results with the exact moments and cross-coordinate dependence
induced by forced-coordinate binomial crossover.

\item
We characterize how \texttt{PRT}, the SOMA path parameter $t$, the
differential weight $F$, and the crossover rate $\mathrm{CR}$ control proposal
scale, dimensionality, covariance, contraction, and orientation.

\item
We use the resulting geometric quantities to construct Geometry-Controlled
SOMA, Rotation-Aware SOMA, and an experimental iSOMA extension combining
rotation-aware masking, success-history adaptation, and linear population-size
reduction. Their performance is evaluated on the complete noiseless BBOB
benchmark across three dimensions and two evaluation budgets.

\item
Exact enumeration and Monte Carlo experiments verify the analytical
identities and quantify the distortions introduced by mask conditioning,
boundary repair, and fitness-based selection.
\end{enumerate}

Section~\ref{sec:algorithmic-background} introduces SOMA and DE and defines the
scope of the canonical variants considered in the analysis. The following
sections develop the SOMA operator factorization and stochastic proposal
geometry and then derive the corresponding finite-population results for DE.
The analytical quantities are subsequently used to construct and evaluate
operator-guided SOMA variants on the noiseless BBOB benchmark. The final
sections discuss the empirical implications, limitations, and directions for
further development. The appendices provide detailed numerical verification
of the analytical identities and complete function-wise convergence profiles.

\section{SOMA and Differential Evolution: Algorithmic Background and Positioning}
\label{sec:algorithmic-background}

\subsection{Self-Organizing Migrating Algorithm}
\label{subsec:soma-background}

SOMA was introduced as a population-based optimizer inspired by the
competitive--cooperative behavior of a group of agents. Its defining mechanism
is migration rather than reproduction in the conventional genetic-algorithm
sense: individuals evaluate a finite set of candidate positions along paths
directed toward selected targets. In the canonical continuous formulation, the
principal controls are the population size, the migration \texttt{Step}, the
\texttt{PathLength}, and the perturbation probability \texttt{PRT}
\cite{davendra2016soma,skanderova2023soma}. The associated binary perturbation
vector determines which coordinates participate in a displacement and hence
restricts an individual proposal to an active coordinate subspace.

Different SOMA strategies primarily modify target construction and update
semantics. In canonical All-to-One, every non-leader individual migrates toward
the current best individual. All-to-All evaluates migrations toward multiple
population members; All-to-Random replaces the deterministic best target by a
sampled member; and cluster-based strategies select targets within locally
identified subpopulations. Discrete, multi-objective, constrained, hybrid, and
parallel variants modify the representation, selection, or implementation
layers while retaining the migration principle
\cite{davendra2016soma,skanderova2023soma}.

A further distinction concerns the temporal use of the perturbation mask. Some
formulations reuse one perturbation vector throughout a complete migration
path, whereas others regenerate the mask at individual path points. A shared
mask couples all path proposals through one random coordinate-selection
operator; regenerated masks produce path-point-specific random operators. The
proposal moments derived later therefore state explicitly whether masks are
independent, reused, or conditioned to contain at least one active coordinate.

The present analysis concerns the canonical continuous, leader-directed,
static-origin proposal rule. Every path proposal is generated from the same
migration-origin point, and the best evaluated path point is selected only after
the path has been sampled. Immediate-update variants instead replace the
migration origin as soon as an improvement is found and therefore generate a
sequential, state-dependent composition of operators. Modern variants can add
group-based target selection, reverse-ordered path evaluation, progressive
search-space narrowing, and stagnation-triggered replacement, as illustrated by
iSOMA \cite{diep2022isoma}. Such mechanisms remain compatible with the
operator--selection viewpoint, but their exact representation requires an
augmented state containing the variables that determine the current target,
bounds, history, and update condition.

Table~\ref{tab:soma-operator-taxonomy} summarizes representative established
SOMA strategies in the terminology used by the operator framework. The three
operator-guided variants proposed in this paper are better introduced later,
together with their definitions and experimental motivation, rather than in the
background taxonomy.

\begin{table*}[htpb]
\centering
\caption{Representative established SOMA strategies interpreted through the
proposal-operator framework. The table is a structural taxonomy rather than an
exhaustive historical catalogue.}
\label{tab:soma-operator-taxonomy}
\footnotesize
\renewcommand{\arraystretch}{1.18}
\begin{tabularx}{\textwidth}{@{}p{0.14\textwidth}p{0.18\textwidth}p{0.20\textwidth}p{0.18\textwidth}X@{}}
\toprule
Strategy or variant & Target construction & Perturbation or search geometry & Update rule & Operator interpretation \\
\midrule
Canonical All-to-One
& Global best individual
& Binary diagonal \texttt{PRT} mask and discretized path
& Best point selected after evaluating the path
& Family of affine proposals sharing one migrant--leader state. \\

All-to-All
& Every other population member
& Binary diagonal \texttt{PRT} mask
& Pairwise path evaluation
& Collection of pair-conditioned affine operators with a changing target. \\

All-to-Random
& Randomly sampled population member
& Binary diagonal \texttt{PRT} mask
& Pathwise selection
& Random-anchor affine operator; target randomness contributes an additional covariance component. \\

Cluster-based SOMA
& Leader selected within an inferred cluster
& Usually coordinate-masked, locally anchored movement
& Cluster-conditioned migration and selection
& Mixture of locally conditioned proposal operators indexed by cluster membership. \\

iSOMA \cite{diep2022isoma}
& Leaders and migrants selected from sampled groups
& Reverse-ordered jumps and progressively narrowed search space
& Immediate improvement update and stagnation-triggered replacement
& State-, bound-, and history-dependent composition requiring an augmented algorithmic state. \\
\bottomrule
\end{tabularx}
\end{table*}

\subsection{Differential Evolution}
\label{subsec:de-background}

DE is a population-based evolutionary optimizer for continuous parameter
spaces. In the canonical \texttt{DE/rand/1/bin} strategy, each population member
acts as a target vector. Three distinct donor indices are sampled, a mutant is
constructed by adding a scaled population difference to a randomly selected
base vector, and binomial crossover combines the mutant with the target. The
standard crossover implementation forces at least one coordinate to be inherited
from the mutant. Finally, greedy one-to-one selection retains the trial vector
when it is no worse than its target \cite{storn1997differential,das2011survey}.
The strategy notation \texttt{DE/x/y/z} records the rule used to choose the base
vector, the number of difference vectors, and the crossover type. Thus,
\texttt{DE/rand/1/bin} uses a random base, one scaled difference, and binomial
crossover.

The behavior of canonical DE is governed mainly by the population size, the
differential weight $F$, the crossover rate $\mathrm{CR}$, and the mutation
strategy. The differential weight scales population-derived directions, while
crossover determines how much of the mutant is transferred to the trial. The
state-of-the-art and updated surveys of DE document the subsequent development
of alternative base-vector rules, multiple-difference strategies, adaptive and
self-adaptive parameter control \cite{brest2006jde,qin2009sade,zhang2009jade}, ensembles of mutation operators, population
size adaptation, hybridization, and extensions to constrained, multi-objective,
large-scale, and uncertain optimization
\cite{das2011survey,das2016recent}. Although these variants differ
substantially at the algorithmic level, many retain the same modular structure:
donor construction, recombination, objective evaluation, and target--trial
selection.

An influential adaptive lineage begins with SHADE, which stores successful
values of the differential weight and crossover rate in historical memories
\cite{tanabe2013shade}. L-SHADE augments this mechanism with linear
population-size reduction \cite{tanabe2014lshade}, while iL-SHADE introduces
further modifications to improve its single-objective real-parameter
performance \cite{brest2016ilshade}. These methods provide the adaptation
background for the DE reference algorithm and for the success-history and
population-reduction components used later in the experimental
iL-SHOMA-RA method.

Theoretical analyses of DE have addressed convergence, invariance, differential
mutation, crossover, population diversity, and population dynamics
\cite{opara2019survey}. Proposal generation is comparatively amenable to
algebraic and probabilistic analysis because, after donor indices and crossover
choices are fixed, the trial is a linear combination of population members.
Greedy replacement is harder to characterize in a problem-independent manner
because the acceptance event depends on the objective function. This separation
motivates the present focus on exact finite-population mutant moments and the
additional covariance and dependence introduced by binomial crossover, while
treating selection and boundary repair as subsequent transformations.

\subsection{Common proposal--selection structure and relation to prior work}
\label{sec:background-positioning}

SOMA and DE differ in how their proposal endpoints are constructed. Canonical
SOMA uses a selected target and samples a discretized, coordinate-masked path
from the migrant toward that target. Canonical \texttt{DE/rand/1/bin} constructs
a random mutant from three sampled population members and then uses a
coordinate mask to combine it with the target vector. Conditional on the
required state and random choices, both proposal mechanisms are affine or
linear; their complete updates remain nonlinear because target selection,
boundary repair, and survivor selection depend on fitness or feasibility.

The scope of the exact derivations is therefore deliberately narrower than a
complete population-dynamics theory. For SOMA, the formulas apply to the raw
continuous, leader-directed, static-origin proposal under the stated mask model.
For DE, they apply to canonical \texttt{DE/rand/1/bin} under ordered donor
sampling without replacement and standard forced-coordinate binomial
crossover. Pairwise, random-target, clustered, adaptive, or immediate-update
variants can be represented conditionally by augmenting the state with the
variables that determine their target, bounds, history, or update event, but
those variants do not share identical unconditional dynamics.

Table~\ref{tab:related-positioning} positions the present analysis relative to
the most closely related studies.

\begin{table}[htpb]
\centering
\caption{Positioning of the present analysis relative to closely related work.}
\label{tab:related-positioning}
\scriptsize
\renewcommand{\arraystretch}{1.18}
\begin{tabularx}{\textwidth}{@{}p{0.19\textwidth}XX@{}}
\toprule
Study &
Primary objective and methodology &
Relation to the present work \\
\midrule

DE surveys
\cite{das2011survey,das2016recent,opara2019survey}
&
Review DE strategies and theoretical results concerning mutation, crossover,
parameter control, convergence, invariance, diversity, and population dynamics.
&
Establish the broader DE context. The present work develops an explicit
augmented-state representation and exact finite-population proposal moments for
canonical \texttt{DE/rand/1/bin}, while treating repair and selection as
separate transformations.
\\

Guo and Yang \cite{guo2015eigenvector};
Caraffini and Neri \cite{caraffini2019rotation}
&
Develop covariance-eigenvector and rotation-invariant crossover mechanisms
for DE and study their behavior under rotated problems.
&
Establish the closest representation-oriented precedent. The present work
applies a population-derived basis to the perturbation mask of the unchanged
leader-directed SOMA proposal and derives its operator interpretation. \\

Opara and Arabas
\cite{opara2018mutation,opara2019contour}
&
Derive distributional properties of differential mutation strategies and
formalize the contour-fitting behavior of DE.
&
Provide the closest probabilistic DE analysis. The present work additionally
uses an augmented linear representation, exact finite-population donor
sampling, and forced-coordinate crossover moments, with repair and selection
treated separately. \\

Tomczak et al. \cite{tomczak2020reversible}
&
Design ADE and RevDE candidate-generation rules by applying invertible linear
transformations to triplets of population members and analyze reversibility and
eigenvalue behavior.
&
Use linear algebra for algorithm synthesis. The present work instead uses
linear algebra to analyze unchanged SOMA and DE rules and extends the
deterministic representation to stochastic proposal geometry.
\\

Pluh\'a\v{c}ek et al. \cite{pluhacek2021explaining}
&
Empirically study the influence of \texttt{PRT} on SOMA trajectories,
parameter-space coverage, population diversity, and convergence.
&
Provide the main empirical SOMA baseline. The present work derives
proposal-level moments and geometric quantities that quantify mask activation
under explicitly stated probabilistic models.
\\

Davendra and Zelinka; Skanderov\'a
\cite{davendra2016soma,skanderova2023soma}
&
Describe the historical development, migration strategies, variants,
applications, and implementation choices of SOMA.
&
Establish the SOMA-family context. The present analysis is restricted to the
canonical continuous proposal and develops exact operator-level stochastic
geometry rather than a general algorithmic survey.
\\

Diep et al. \cite{diep2022isoma}
&
Introduce iSOMA with group-based organization, reverse-ordered jumps,
immediate updating, narrowed search bounds, stagnation handling, and a drone
path-planning application.
&
Represent a modern state-dependent SOMA variant. The present framework
identifies the additional state variables required to express such a variant as
a conditional composition of proposal and selection operators.
\\

Present work
&
Factorizes canonical SOMA and \texttt{DE/rand/1/bin} into raw proposal, repair,
and fitness-dependent selection stages and derives exact operator forms and
stochastic moments.
&
Combines conditional linear representation, finite-population sampling,
random-mask analysis, and explicit examination of distortions introduced by
repair and selection.
\\

\bottomrule
\end{tabularx}
\end{table}

% Relocation recommendation:
% Move the GC-SOMA, RA-SOMA, and SH-SOMA rows from the original SOMA taxonomy
% table to the beginning of the later ``Operator-Guided SOMA Design'' section,
% after those variants have been motivated and named. This avoids introducing
% original methods inside the background/history subsection.

\section{Operator--Selection Factorization of SOMA}
\label{sec:SOMA}

Consider the minimization of an objective function
$f:\Omega\subseteq\mathbb{R}^{D}\rightarrow\mathbb{R}$. At migration cycle
$m$, SOMA maintains a population
$\mathcal{X}^{(m)}=\{\bm{x}^{(m)}_1,\ldots,\bm{x}^{(m)}_N\}$ and selects a
leader according to
\begin{equation}
\bm{L}^{(m)}
\in
\operatorname*{arg\,min}{\bm{x}\in\bm{X}^{(m)}} f(\bm{x}).
\label{eq}
\end{equation}
The leader-selection mechanism may vary across SOMA strategies, but it is an
objective-dependent operation. To analyze the variation stage, fix one
migration cycle and one migrant and write the migration origin as $\bm{x}$ and
the selected target as $\bm{L}$.

Let $t_1,\ldots,t_J$ denote the path parameters determined by
\texttt{Step} and \texttt{PathLength}, and let
$\Pi_j=\operatorname{diag}(\pi_{j,1},\ldots,\pi_{j,D})$ be the perturbation
mask at path point $j$. The raw proposal is
\begin{equation}
\bm{z}_j
=
\bm{x}
+
t_j\bm{\Pi}_j(\bm{L}-\bm{x}).
\label{eq1}
\end{equation}
The algorithmic meaning of the path, mask, and static-origin assumption was
specified in Section~\ref{subsec:soma-background}. The purpose of the present
section is to separate this proposal equation from repair and pathwise survivor
selection and to derive its exact affine and augmented linear representations.

\begin{figure}[htpb]
    \centering
    \includegraphics[width=0.65\linewidth]{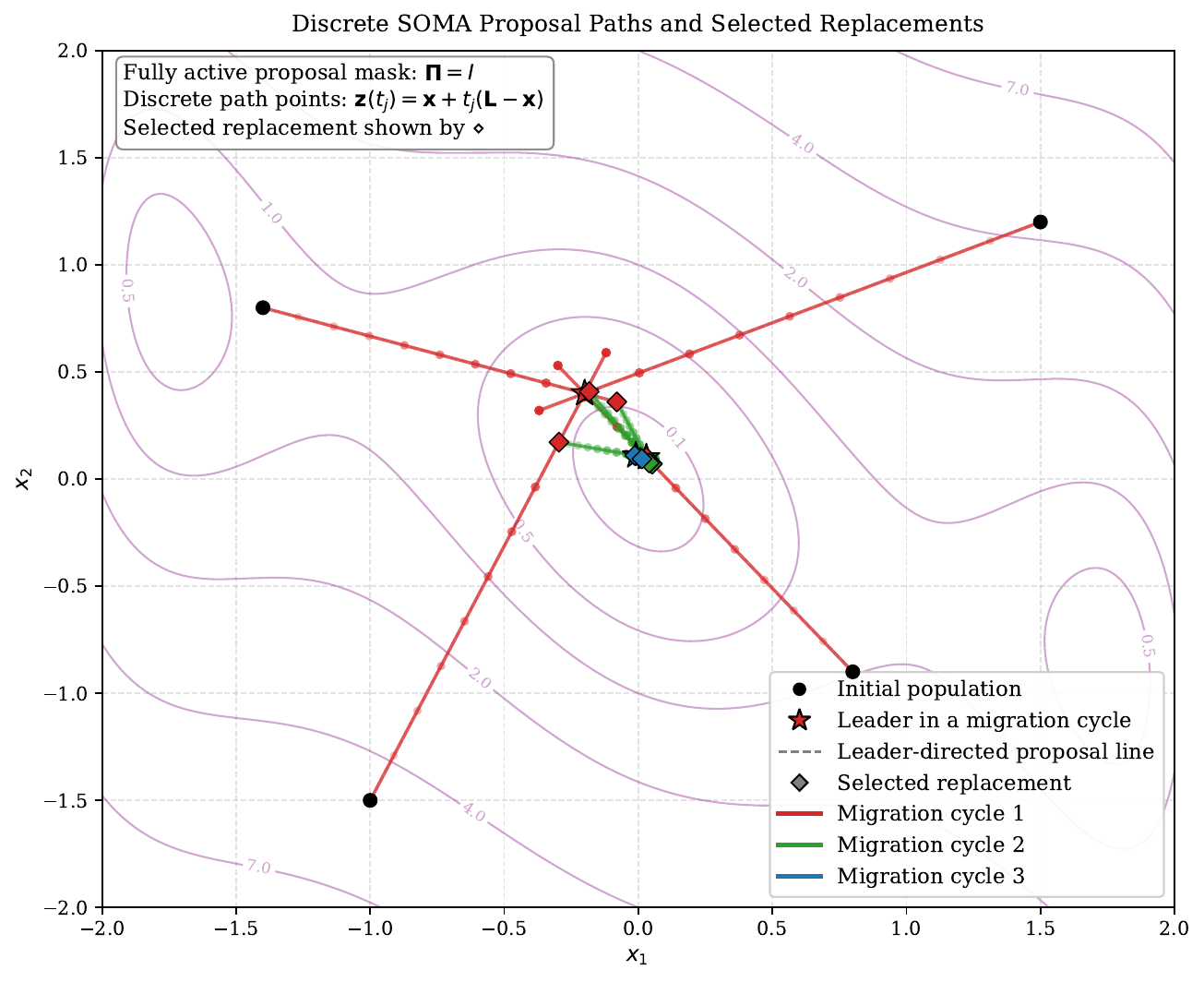}
    \caption{
Discrete SOMA proposal paths and selected replacements on the
Three-Hump Camel function for three successive migration cycles.
For each non-leader individual, candidate points are generated along a
leader-directed linear path using a fully active perturbation mask
$\bm{\Pi}=I$, so that the path points satisfy
$\bm{z}(t_j)=\bm{x}+t_j(\bm{L}-\bm{x})$.
The colored circular markers show the discretized proposal points
associated with the path parameters $t_j$, and the dashed segments
indicate the corresponding leader-directed proposal lines.
}
\label{fig:soma_linear_paths}
\end{figure}

Figure~\ref{fig:soma_linear_paths} visualizes the deterministic
fully-active case $\bm{\Pi}=I$, in which each SOMA proposal path reduces
to a discretized line segment from the migrant toward the current
leader. This figure is intended to illustrate the geometric structure of
the proposal rule. Stochastic coordinate masking and its induced
proposal distribution are examined separately in
Figure~\ref{fig:soma_prt_geometry}.

The perturbation mask determines which coordinates participate in the movement and serves the purpose of exploration. If $\pi_{j,d}=1$, coordinate $d$ is displaced along the migrant--leader direction. If $\pi_{j,d}=0$, that coordinate remains equal to its value in the migration-origin point.

The path parameter controls the geometry of the active coordinates. Values $0<t_j<1$ interpolate between the migrant and the leader, $t_j=1$ copies the active leader coordinates exactly, and $t_j>1$ generates an overshooting proposal beyond the leader. This overshooting mechanism allows SOMA to investigate regions that are not contained in the line segment joining the two individuals.

Although the proposals are indexed by $j$, they should not generally be interpreted as a recursively generated trajectory. In the standard SOMA path construction, every proposal $\bm{z}_j$ is obtained from the same migration-origin point $\bm{x}$. In particular, $\bm{z}_{j+1}$ is not normally generated by applying the next path transformation to $\bm{z}_j$. The term \emph{linear proposal path} therefore refers to a discrete collection of geometrically ordered proposals, each produced by an operator acting on the same migrant--leader state.

This static-origin interpretation applies to the canonical pathwise-update
formulation analyzed in this paper. Adaptive or immediate-update variants may
replace the migration origin as soon as an improving point is found, in which
case the migration becomes a genuinely sequential, state-dependent composition
of proposal operators rather than a set of proposals sharing one fixed origin.

After the proposals have been evaluated, the next position of the migrant is selected from
\begin{equation}
\mathcal{C}
=
{\bm{x},\bm{z}_1,\ldots,\bm{z}_J},
\qquad
\bm{x}^{+}
\in
\operatorname*{arg\,min}{\bm{y}\in\mathcal{C}} f(\bm{y}).
\label{eq2}
\end{equation}
Including the original migrant in $\mathcal{C}$ ensures that the selected position is not worse than the migration origin under exact objective evaluation.

Equations~\eqref{eq1} and
\eqref{eq2} describe two mathematically different operations. The first is a parameterized geometric proposal rule. The second is an objective-dependent decision rule involving fitness evaluation and an $\arg\min$ operation. Consequently, the proposal rule may possess an exact linear representation even though the complete migration remains nonlinear.

\subsection{Exact linear representation of the proposal rule}

For a fixed path parameter $t$ and mask $\bm{\Pi}$, define the augmented migrant--leader state
\[
\bm{s}
=
\begin{bmatrix}
\bm{x}\\
\bm{L}
\end{bmatrix}
\in\mathbb{R}^{2D}.
\]
The proposal rule can then be written as
\begin{equation}
\begin{bmatrix}
\bm{z}\\
\bm{L}
\end{bmatrix}
=
\underbrace{
\begin{bmatrix}
I-t\bm{\Pi} & t\bm{\Pi}\\
0 & I
\end{bmatrix}}_{\displaystyle
\mathcal{M}(t,\bm{\Pi})}
\begin{bmatrix}
\bm{x}\\
\bm{L}
\end{bmatrix}.
\label{eq3}
\end{equation}

In the original migrant coordinates, Eq.~\eqref{eq1} is affine because the leader acts as a translation anchor. After augmenting the state with the leader position, the same proposal becomes an exact linear matrix--vector product.

The coefficients of $\mathcal{M}(t,\bm{\Pi})$ depend only on the path parameter and perturbation mask. They do not depend on the numerical values of $\bm{x}$ or $\bm{L}$, and they contain no objective value, gradient, rank, or fitness-conditioned switch. Once the migrant, leader, path parameter, and mask have been specified, the proposal is therefore produced by an objective-independent linear operator.

This statement does not imply that the proposal itself is independent of previous objective information. The leader appearing in the input state was selected using fitness values, and the migrant is the result of previous selection decisions. The exact decoupling concerns the \emph{coefficients of the proposal operator}: objective information enters through the current state, but not through the matrix that maps this state to a path proposal.

\begin{figure}[htpb]
\centering
\resizebox{\linewidth}{!}{
\begin{tikzpicture}[
    node distance=1.5cm,
    auto,
    block/.style={
        draw,
        rounded corners,
        inner sep=10pt,
        align=center,
        minimum height=2em,
        font=\sffamily\small
    },
    workflow/.style={
        block,
        minimum width=8em
    },
    category/.style={
        draw,
        rounded corners,
        inner sep=10pt,
        align=center,
        minimum height=3.5em,
        minimum width=12em,
        font=\sffamily\large
    },
    arrow/.style={
        -Latex,
        ultra thick
    }
]

% Top Workflow Row Nodes
\node [workflow, draw=black] (combined) {
    Current migrant-leader state $(\bm{x},\bm{L})$ \\
    from population at migration $X^{(m)}$ \\
};

\node [workflow, draw=black, dashed, right=of combined, node distance=2.5cm] (path) {
    Linear proposal path \\
    $\mathcal{M}(t_j, \mathbf{\Pi}_j)$
};

\node [workflow, draw=black, dashed, right=of path, node distance=2.5cm] (eval) {
    Evaluate candidates \\
    and retain best
};

\node [workflow, draw=black, right=of eval, node distance=2.5cm] (update) {
    Updated migrant / \\
    next population state
};

% Bottom Category Row Nodes
\node [category, below=of path, node distance=2cm] (variation) {
    \textbf{Variation} \\
    exact linear generation \\
    of path candidates
};

\node [category, below=of eval, node distance=2cm] (selection) {
    \textbf{Selection} \\
    survival / replacement \\
    under fitness
};

% Set Description Position (Midpoint between Path and Eval, slightly high)
\coordinate (mid_path_eval) at ($(path.east)!0.5!(eval.west)$);
\node [font=\sffamily\small, above=of mid_path_eval, node distance=0.1cm] (setdesc) {
    $\mathcal{C}^{(m)} = \{\mathbf{x}, \mathbf{z}_1, \dots, \mathbf{z}_J\}$
};

% Workflow Arrows
\draw [arrow] (combined.east) -- (path.west);
\draw [arrow] (path.east) -- (eval.west);
\draw [arrow] (eval.east) -- (update.west);

% Vertical category lines
\draw [very thick] (variation.north) -- (path.south);
\draw [very thick] (selection.north) -- (eval.south);

\end{tikzpicture}
}
\caption{
Variation--selection decomposition of one SOMA migration.
The current population first undergoes \emph{selection} to identify the
leader that defines the search direction. Conditional on the migrant,
leader, path parameter, and perturbation mask, the candidate points along
the path are then generated by an exact objective-independent linear
\emph{variation} operator. The resulting candidate set is finally subjected
to \emph{selection}, where objective-function evaluations determine which
candidate survives into the next population state. This separates the
exactly representable proposal mechanism from the objective-dependent
selection pressure exerted before and after variation.
}
\label{fig:variation_selection_soma}
\end{figure}
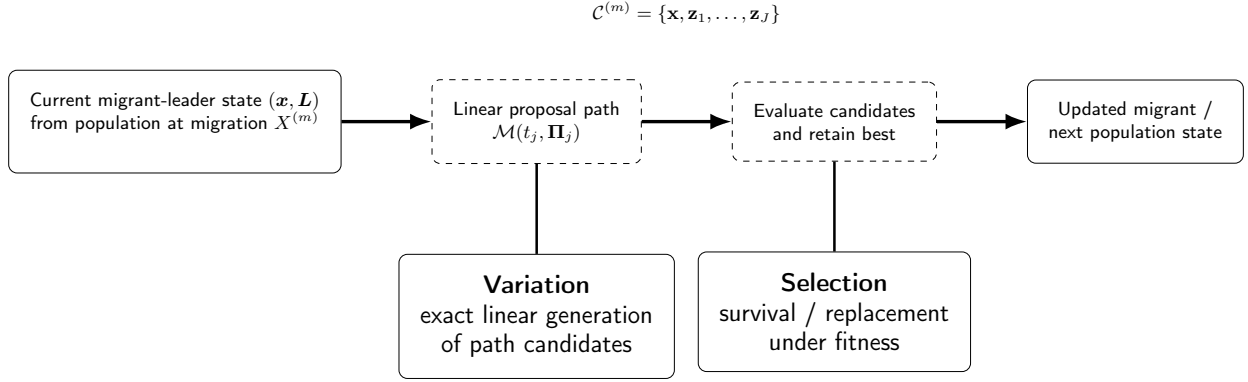

At the population level, one SOMA migration follows the standard
variation--selection logic of evolutionary computation. Fitness-based
selection first determines the leader that guides the search. The
resulting migrant--leader pair then generates a path of candidate
solutions through an exact linear variation operator. Finally, the
generated candidates are evaluated and selection determines which point
survives into the next population state. 
Figure~\ref{fig:variation_selection_soma}
summarizes this decomposition.

The decomposition is also consistent with the variation--selection
organization underlying evolutionary computation
\cite{eiben2015introduction, doerr2021survey}. In the biological interpretation associated
with the Modern Evolutionary Synthesis, mechanisms that generate
heritable variation are conceptually distinguished from differential
survival and reproduction under selection
\cite{huxley1942evolution}. Evolutionary algorithms retain this abstract
organization without reproducing the details of biological evolution:
variation constructs candidate solutions, whereas fitness-based
selection changes their representation in the subsequent population.

Within this interpretation, the SOMA proposal path is a structured
leader-guided variation mechanism, and the comparison of its path points
exerts selection pressure. The contribution of the present analysis is
to show that the internal variation mechanism can possess an exact linear
operator form even though the surrounding selection process is nonlinear.
The biological analogy concerns this separation of roles; it does not
imply that biological variation itself is linear or that SOMA constitutes
a literal model of biological evolution.

\subsection{Extension to random, pairwise, and state-dependent targets}
\label{subsec:soma-target-extension}

The canonical expression can be embedded in a broader class of
state-conditioned anchored affine proposals,
\begin{equation}
\bm z
=
\bm x
+
\bm A_{\theta}(\bm X,\bm H,\omega)
\bigl(
\bm L_{\theta}(\bm X,\bm H,\omega)-\bm x
\bigr),
\label{eq:general-soma-operator}
\end{equation}
where \(\bm X\) is the current population, \(\bm H\) denotes optional
algorithmic history, \(\omega\) collects random auxiliary choices,
\(\bm L_{\theta}\) is a target-construction rule, and \(\bm A_{\theta}\)
is the proposal operator. Canonical SOMA is recovered by
\(\bm A_{\theta}=t\bm\Pi\) and by taking \(\bm L_{\theta}\) as the current
leader. All-to-All uses the same conditional operator with a pair-specific
target; All-to-Random makes the target itself random; cluster-based methods
condition it on cluster membership; and adaptive variants allow both terms to
depend on \(\bm H\).

For a random target \(\bm L\), the conditional moments derived below remain
valid given \(\bm L\). The unconditional proposal covariance then satisfies
\begin{equation}
\operatorname{Cov}(\bm z)
=
\mathbb{E}_{\bm L}
\left[
\operatorname{Cov}(\bm z\mid\bm L)
\right]
+
\operatorname{Cov}_{\bm L}
\left(
\mathbb{E}[\bm z\mid\bm L]
\right),
\label{eq:random-target-total-covariance}
\end{equation}
which separates mask-induced uncertainty from uncertainty caused by target
selection. Equation~\eqref{eq:general-soma-operator} is not used to claim that
all SOMA variants share identical dynamics. Rather, it identifies the minimum
additional conditioning needed to transfer the present proposal-level analysis
to pairwise, clustered, adaptive, or bounded-search variants.

\section{Geometry of the SOMA Operator}
\label{sec:2}

The SOMA proposal becomes especially transparent when it is written
relative to the current leader. Define the leader-relative displacement
\[
\bm{e}=\bm{x}-\bm{L}.
\]
Since the proposal rule is
\[
\bm{z}=\bm{x}+t\bm{\Pi}(\bm{L}-\bm{x}),
\]
the corresponding proposal displacement relative to the leader is
\begin{equation}
\bm{e}^{+}=\bm{z}-\bm{L}=(I-t\bm{\Pi})\bm{e}.
\label{eq:soma_relative}
\end{equation}

Equation~\eqref{eq:soma_relative} is the key geometric form of the SOMA
operator. It shows that the proposal mechanism does not require a more
complicated nonlinear interpretation: relative to the leader, SOMA simply
rescales the current displacement coordinate-wise. Inactive coordinates
remain unchanged, while active coordinates are multiplied by the factor
\(1-t\).

For an active coordinate, Eq.~\eqref{eq:soma_relative} reduces to
\[
e_d^{+}=(1-t)e_d.
\]
The path parameter \(t\) therefore has an immediate geometric meaning.
It determines whether the proposal interpolates toward the leader,
projects onto the leader coordinate, or overshoots beyond it.

The set
\[
\mathcal{E}_{0}
=
\{
(\bm{x},\bm{L})\in\mathbb{R}^{2D}:\bm{x}=\bm{L}
\}
\]
forms a consensus manifold of the proposal rule: if the migrant already
coincides with the leader, then the proposal leaves this state unchanged.
This is a geometric property of the proposal operator itself and should
not be interpreted as a convergence theorem for the complete SOMA
optimizer.

\begin{table}[htpb]
\centering
\caption{
Coordinate-wise geometry of an active SOMA proposal.
The quantities are expressed relative to the current leader.
}
\label{tab:soma_geometry}
\begin{tabular}{lll}
\hline
Path parameter & Multiplier \(1-t\) & Geometric effect\\
\hline
\(t=0\) & \(1\) & no movement\\
\(0<t<1\) & \((0,1)\) & interpolation toward the leader\\
\(t=1\) & \(0\) & projection onto the leader coordinate\\
\(1<t<2\) & \((-1,0)\) & overshooting with reduced distance\\
\(t=2\) & \(-1\) & reflection with unchanged distance\\
\(t>2\) & less than \(-1\) & overshooting with increased distance\\
\hline
\end{tabular}
\end{table}

Table~\ref{tab:soma_geometry} shows that overshooting is not necessarily
expansive. For \(1<t<2\), the proposal crosses the leader but remains
closer to it than the migration origin. Only values \(t>2\) increase the
magnitude of an active leader-relative coordinate.

If \(K=\operatorname{tr}(\bm{\Pi})\) denotes the number of active
coordinates, then the proposal rescales \(K\) leader-relative directions
by the common factor \(1-t\). Consequently, the volume change associated
with one proposal operator is
\begin{equation}
\det \mathcal{M}(t,\bm{\Pi})=(1-t)^K.
\label{eq:soma_det}
\end{equation}
For \(0<t<2\), every active relative coordinate contracts in magnitude.
At \(t=1\), the operator becomes singular whenever at least one coordinate
is active, because those active directions are mapped exactly onto the
leader. For \(t>2\), the active relative directions expand.

The determinant in Eq.~\eqref{eq:soma_det} describes the deformation
induced by a single proposal operator. It should not be interpreted as
repeated contraction of the standard SOMA trajectory, because the path
points are generated from a common migration origin rather than by
iterative composition of the same state update.

\subsection{Bernoulli perturbation masks}
\label{subsec:bernoulli_masks}

The perturbation mask can be interpreted as a random
coordinate-selection operator. Assume that its diagonal entries are
independent Bernoulli variables,
\[
    \pi_d\sim\operatorname{Bernoulli}(p),
\]
where \(p\in[0,1]\) corresponds to the SOMA perturbation parameter
\texttt{PRT} under the independent-mask model
\cite{davendra2016soma}. For the migrant--leader displacement
\(\bm r=\bm L-\bm x\), the random proposal is
\[
    \bm z=\bm x+t\bm\Pi\bm r.
\]

Its conditional first and second moments are
\begin{align}
    \mathbb E[\bm z\mid\bm x,\bm L,t]
    &=
    \bm x+pt\bm r,
    \label{eq:proposal_mean}\\
    \operatorname{Cov}(\bm z\mid\bm x,\bm L,t)
    &=
    p(1-p)t^2
    \operatorname{diag}
    \left(r_1^2,\ldots,r_D^2\right).
    \label{eq:proposal_covariance}
\end{align}

The expected proposal therefore lies on the migrant--leader line, with
\(pt\) acting as an effective path parameter. The covariance is diagonal
because the mask coordinates are independent, and its variance in
coordinate \(d\) is proportional to \(r_d^2\). Hence the proposal
uncertainty is largest in coordinates where the migrant and leader differ
most, while coordinates in which they agree have zero variance.

This induced covariance should not be confused with the covariance
maintained by CMA-ES \cite{hansen2001cmaes}. CMA-ES explicitly learns and
updates a generally dense sampling covariance from selected search steps.
SOMA stores no covariance matrix; Eq.~\eqref{eq:proposal_covariance} is
only a statistical consequence of its existing masked proposal rule.

A compact measure of leader-directed movement is the normalized expected
squared distance
\begin{equation}
    \frac{
        \mathbb E[
            \|\bm z-\bm L\|^2
            \mid\bm x,\bm L,t
        ]
    }{
        \|\bm x-\bm L\|^2
    }
    =
    1-pt(2-t).
    \label{eq:expected_leader_distance}
\end{equation}
For \(p>0\), this quantity is smaller than one when \(0<t<2\), equals
one at \(t=2\), and exceeds one for \(t>2\). Thus, overshooting with
\(1<t<2\) still reduces the expected distance from the leader. The
expected squared step length from the migrant is, similarly,
\(pt^2\|\bm r\|^2\).

\begin{figure}[htpb]
    \centering
    \includegraphics[width=\linewidth]
    {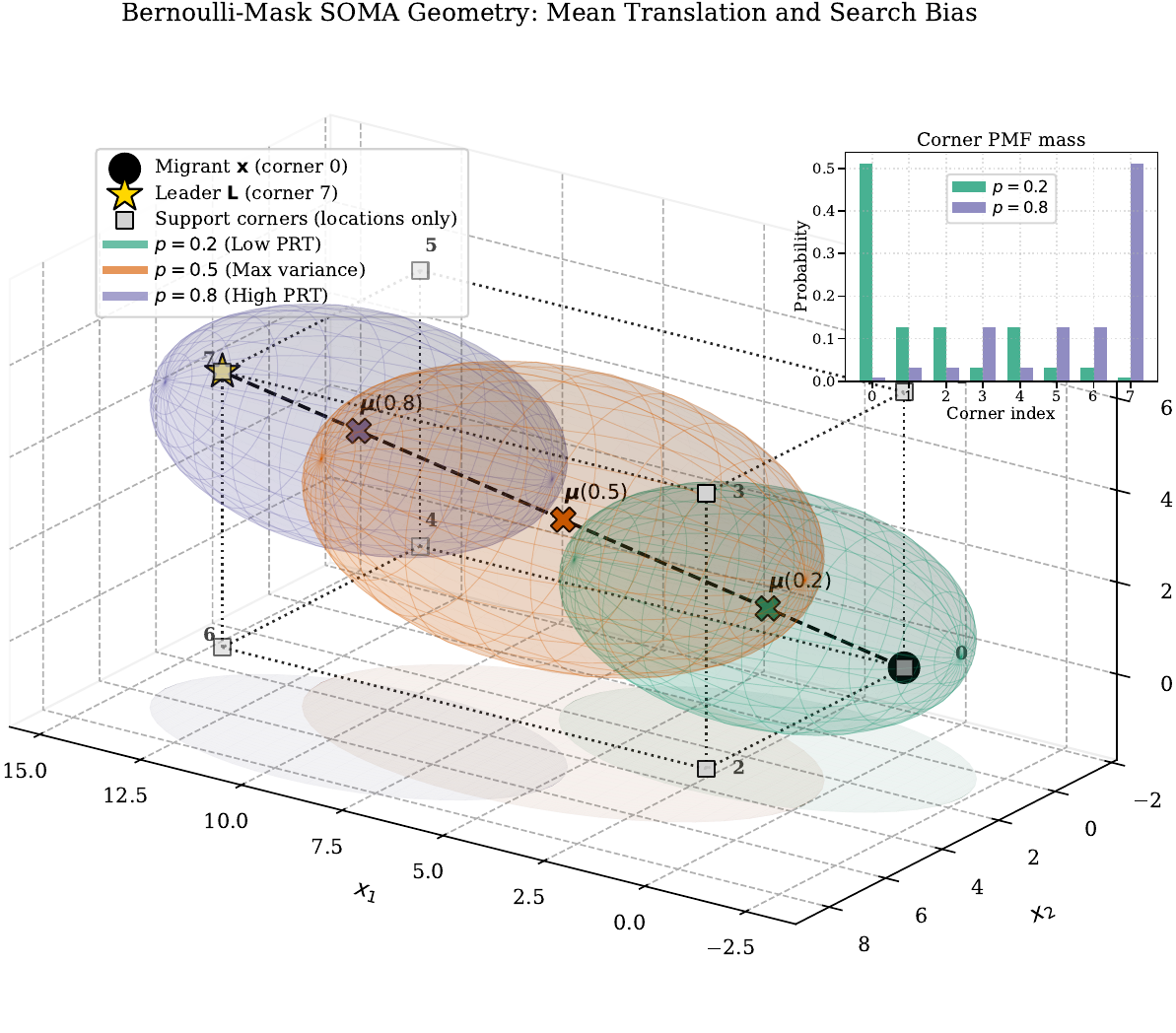}
    \caption{
    Bernoulli-mask SOMA proposal geometry for fixed migrant
    \(\bm x\), leader \(\bm L\), and path parameter \(t\).
    The gray markers show the exact mask-generated support, while the
    colored markers and ellipsoids show the proposal means and covariance
    summaries for three values of \(p=\texttt{PRT}\). Changing \(p\)
    redistributes probability over the same support, translates the mean
    along the migrant--leader direction, and scales the covariance by
    \(\sqrt{p(1-p)}\). The inset compares the corner probability masses
    for \(p=0.2\) and \(p=0.8\). The ellipsoids are second-moment
    summaries rather than support boundaries or Gaussian confidence
    regions.
    }
    \label{fig:soma_prt_geometry}
\end{figure}

The covariance scale is governed by \(p(1-p)\), which is maximal at
\(p=1/2\). Increasing \texttt{PRT} therefore does not monotonically
increase mask-induced variability: small \(p\) activates few coordinates,
intermediate \(p\) produces the largest variance, and \(p=1\) eliminates
mask randomness.

The number of active coordinates satisfies
\begin{equation}
    K=\operatorname{tr}(\bm\Pi)
    \sim\operatorname{Binomial}(D,p),
    \qquad
    \mathbb E[K]=Dp,
    \qquad
    \Pr(K=0)=(1-p)^D.
    \label{eq:active_coordinate_distribution}
\end{equation}
Thus, \texttt{PRT} also controls the expected dimensionality of an
individual proposal. Implementations that reject all-zero masks instead
use the corresponding distribution conditioned on \(K>0\).

If independent masks are regenerated at \(J\) path points, coordinate
coverage is summarized by
\begin{equation}
\begin{aligned}
    \Pr(\text{one coordinate is covered})
        &=1-(1-p)^J,\\
    \mathbb E[\text{uncovered coordinates}]
        &=D(1-p)^J,\\
    \Pr(\text{all coordinates are covered})
        &=
        \left[1-(1-p)^J\right]^D.
\end{aligned}
\label{eq:path_coverage}
\end{equation}
These expressions do not apply when a single mask is reused throughout
the path or when coordinate activations are correlated.

Figure~\ref{fig:soma_prt_geometry} shows that the support of the
three-dimensional proposal distribution consists of the eight corners
of the axis-aligned box between the migrant and the fully active
proposal. For \(0<p<1\), changing \texttt{PRT} leaves these locations
unchanged but alters their probability masses, the mean, and the
covariance. The cases \(p\) and \(1-p\) produce ellipsoids of equal size
because they share the same factor \(p(1-p)\), although their centers
differ. The ellipsoids remain aligned with the Cartesian axes because
the independent Bernoulli mask introduces no cross-coordinate covariance.

\section{Linear Operator Structure of Differential Evolution}
\label{sec:de-operator}

Using the \texttt{DE/rand/1/bin} notation introduced in
Section~\ref{subsec:de-background}, let $\bm{x}_i$ be the target vector and let
$r_1,r_2,r_3$ be distinct sampled donor indices, also distinct from $i$. The
mutant and trial vectors are
\begin{equation}
\bm{v}=\bm{x}_{r_1}+F(\bm{x}_{r_2}-\bm{x}_{r_3}),
\qquad
\bm{u}_i=(I-C)\bm{x}_i+C\bm{v},
\label{eq:de_combined_proposal}
\end{equation}
where $C$ is the diagonal binomial-crossover mask. The background section
introduced the procedural roles of mutation, crossover, and greedy replacement;
here the same operations are represented as conditional linear maps and their
finite-population proposal moments are derived.

Conditional on the sampled parent indices, crossover mask, and value of
$F$, Eq.~\eqref{eq:de_combined_proposal} is an exact
objective-independent linear transformation of the stacked population
state. For example,
\begin{equation}
\begin{bmatrix}
\bm{u}\cdot i\
\bm{x}\cdot {r_1}\\
\bm{x}\cdot {r_2}\\
\bm{x}\cdot {r_3}
\end{bmatrix}
=
\underbrace{
\begin{bmatrix}
I-\bm{C} & \bm{C} & F\bm{C} & -F\bm{C}\\
0&I&0&0\\
0&0&I&0\\
0&0&0&I
\end{bmatrix}}\cdot {\displaystyle \mathcal{M}\cdot {\mathrm{DE}}}
\begin{bmatrix}
\bm{x}\cdot i\\
\bm{x}\cdot {r_1}\\
\bm{x}\cdot {r_2}\\
\bm{x}\cdot {r_3}
\end{bmatrix}.
\label{eq:de_augmented_operator}
\end{equation}
The lower block rows merely preserve the sampled donor vectors, in the
same way that the SOMA operator preserves the leader.

The similarity between DE and SOMA becomes even more explicit in
leader-directed DE strategies. For example, the
\texttt{DE/current-to-best/1/bin} proposal can be written as
\begin{equation}
\bm{u}\cdot i
=
\bm{x}\cdot i
+
\bm{C}
\left[
F(\bm{x}\cdot {\mathrm{best}}-\bm{x}\cdot i)
+
F(\bm{x}\cdot {r_1}-\bm{x}\cdot {r_2})
\right].
\label{eq:de_current_best}
\end{equation}
The first difference term in Eq.~\eqref{eq:de_current_best} has exactly
the same leader-attraction structure as the SOMA proposal
$t\bm{\Pi}(\bm{L}-\bm{x})$, with the correspondence
$t=F$, $\bm{\Pi}=\bm{C}$, and
$\bm{L}=\bm{x}_{\mathrm{best}}$. DE adds a second population-difference
term that introduces a direction derived from the current population
distribution. SOMA and leader-directed DE can therefore be viewed as
members of a broader class of masked affine difference operators, with DE
combining directed attraction and differential exploration.

The complete DE update is nevertheless nonlinear because the trial vector
is accepted through greedy fitness-based replacement:
\begin{equation}
\bm{x}_i^{+}
=
\begin{cases}
\bm{u}_i, & f(\bm{u}_i)\leq f(\bm{x}_i),\\
\bm{x}_i, & \text{otherwise}.
\end{cases}
\label{eq:de_selection}
\end{equation}
Thus, DE possesses an exact linear variation operator conditional on its
sampled indices and crossover mask, followed by a nonlinear selection
operator.

\subsection{Stochastic proposal geometry of Differential Evolution}
\label{subsec:de_geometry}

The similarity between SOMA and Differential Evolution extends beyond
their deterministic operator forms. Consider the
\texttt{DE/rand/1/bin} trial vector
\begin{equation}
    \bm{u}_i
    =
    \bm{x}_i+\bm{C}(\bm{v}-\bm{x}_i),
    \qquad
    \bm{v}
    =
    \bm{x}_{r_1}
    +
    F(\bm{x}_{r_2}-\bm{x}_{r_3}),
    \label{eq:de_stochastic_trial}
\end{equation}
where $\bm{C}$ is the diagonal binomial-crossover mask. If its
coordinates are initially modeled as independent Bernoulli variables
with activation probability $q=\mathrm{CR}$, then, conditional on the
target and mutant vectors,
\begin{align}
    \mathbb{E}[\bm{u}_i\mid\bm{v}]
    &=
    \bm{x}_i+q(\bm{v}-\bm{x}_i),
    \label{eq:de_conditional_mean}\\
    \operatorname{Cov}(\bm{u}_i\mid\bm{v})
    &=
    q(1-q)
    \operatorname{diag}
    \left(
        (\bm{v}-\bm{x}_i)^{2}
    \right).
    \label{eq:de_conditional_covariance}
\end{align}

Conditional on $\bm{v}$, the DE crossover geometry is therefore
equivalent to the Bernoulli-mask SOMA geometry with the mutant acting
as the fully active endpoint. The exact support contains at most
$2^D$ corners of the axis-aligned box joining $\bm{x}_i$ and $\bm{v}$.
Changing $\mathrm{CR}$ redistributes probability over this support and
translates the conditional mean, but does not alter the support itself
for $0<\mathrm{CR}<1$.

\begin{figure}[htpb]
    \centering
    \includegraphics[width=1\linewidth]{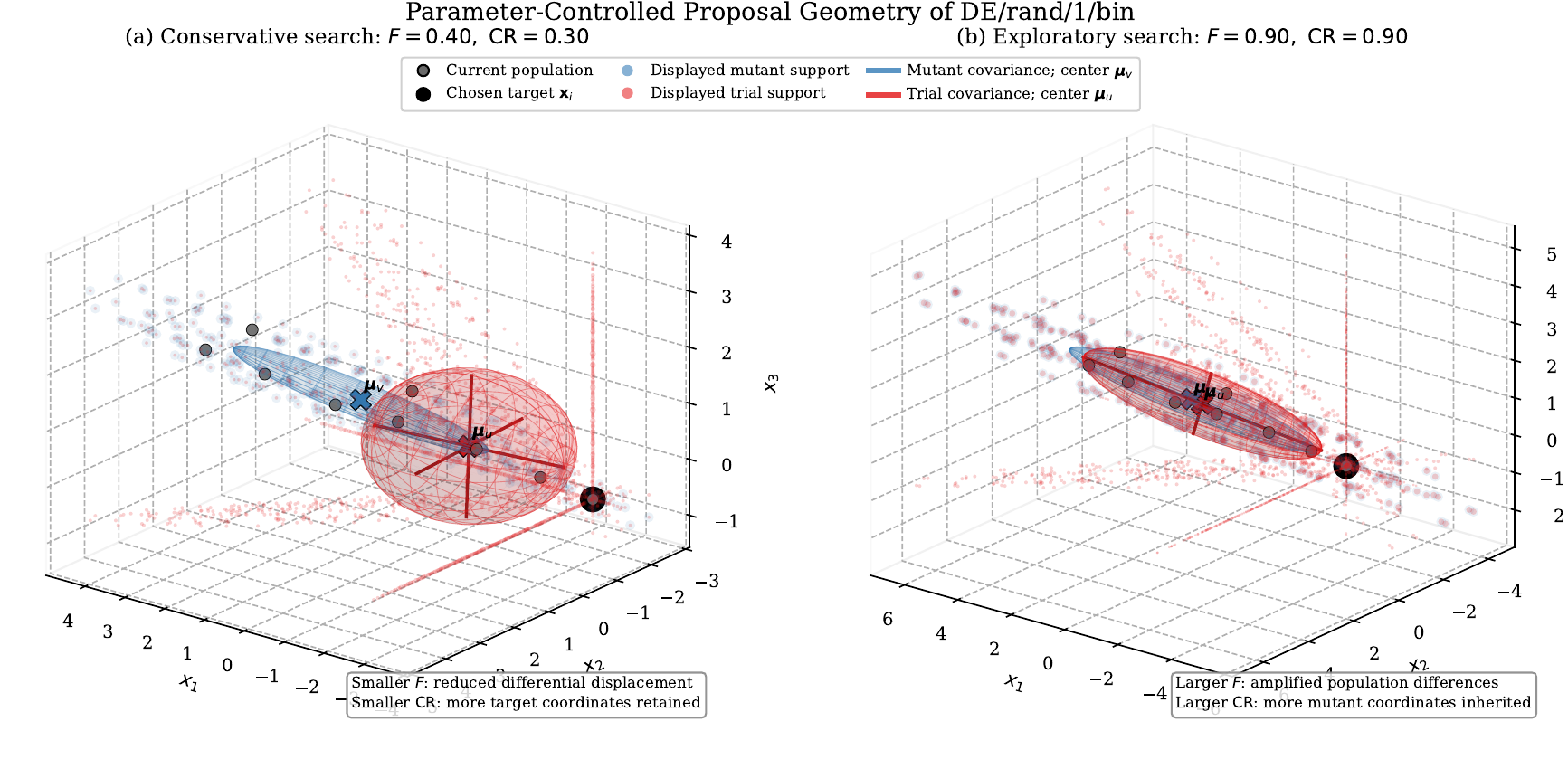}
\caption{
Influence of the differential weight $F$ and crossover rate
$\mathrm{CR}$ on the donor-marginalized proposal geometry of
\texttt{DE/rand/1/bin}. Both panels use the same population and target
vector $\bm{x}_i$. Gray markers denote the current population, blue
points represent possible mutant vectors
$\bm{v}=\bm{x}_{r_1}+F(\bm{x}_{r_2}-\bm{x}_{r_3})$, and red points
represent the corresponding trial vectors after standard
forced-coordinate binomial crossover. The blue and red ellipsoids
summarize the mutant and trial covariance, with centers
$\bm{\mu}_v=\mathbb{E}[\bm{v}]$ and
$\bm{\mu}_u=\mathbb{E}[\bm{u}_i]$, respectively.
(a) Smaller $F$ reduces differential displacement, while smaller
$\mathrm{CR}$ retains more target coordinates and concentrates the
trial distribution near $\bm{x}_i$.
(b) Larger $F$ amplifies population differences, while larger
$\mathrm{CR}$ transfers more mutant coordinates into the trial,
producing a broader distribution that more closely follows the mutant
geometry. The covariance ellipsoids are second-moment summaries of the
finite proposal distributions rather than Gaussian confidence regions.
}
\label{fig:de_parameter_geometry}
\end{figure}

The principal difference is that the DE endpoint $\bm{v}$ is itself
random because it is constructed from sampled population members. Let
\[
    \bm{m}_v=\mathbb{E}[\bm{v}],
    \qquad
    \bm{\Sigma}_v=\operatorname{Cov}(\bm{v})
\]
denote its mean and covariance under donor sampling. Applying the law
of total covariance gives
\begin{equation}
    \operatorname{Cov}(\bm{u}_i)
    =
    q^2\bm{\Sigma}_v
    +
    q(1-q)
    \operatorname{diag}
    \left[
        \operatorname{diag}(\bm{\Sigma}_v)
        +
        (\bm{m}_v-\bm{x}_i)^{2}
    \right].
    \label{eq:de_total_covariance}
\end{equation}
The first term represents uncertainty induced by mutation and donor
selection, whereas the second is the coordinate-aligned covariance
introduced by binomial crossover. Unlike the conditional crossover
covariance, the mutation-induced component may be dense and may rotate
the principal axes of the total trial distribution.

Standard binomial crossover additionally forces one randomly selected
coordinate to be inherited from the mutant. The numerical visualization
therefore enumerates this forced coordinate explicitly rather than
treating all crossover-mask entries as independent.

\section{Operator-Guided SOMA Design and Evaluation}
\label{sec:operator-guided-soma}

\subsection{Algorithms and experimental protocol}
\label{subsec:algorithms-protocol}

The preceding analysis provides quantities that can be used directly for
algorithm design. In particular, the perturbation probability controls the
expected active dimensionality, its interaction with the path parameter
determines the expected contraction toward the leader, and the diagonal
proposal covariance reveals an explicit dependence on the coordinate system.
We use these observations to construct Geometry-Controlled SOMA and
Rotation-Aware SOMA. We additionally combine rotation-aware masking with
success-history adaptation and population-size reduction in an experimental
extension of iSOMA.

For a migrant \(x\), leader \(L\), path parameter \(t\), and Bernoulli mask
\(m\in\{0,1\}^{D}\), the standard proposal is
\begin{equation}
z=x+t\,m\cdot(L-x).
\label{eq:design-standard-soma}
\end{equation}

\paragraph{Geometry-Controlled SOMA.}
Geometry-Controlled SOMA (GC-SOMA) replaces direct specification of the
perturbation probability \(p\) and path parameter \(t\) by an expected active
dimension \(k^\star\) and a desired normalized squared leader distance
\(\rho^\star\). From the proposal moments derived in
Section~\ref{sec:2},
\begin{equation}
p=\frac{k^\star}{D},
\qquad
\rho^\star
=
1-p\,t(2-t),
\qquad
t
=
1\pm
\sqrt{1-\frac{1-\rho^\star}{p}}, 
\qquad
0 < p \le 1, \quad 1 - p \le \rho^* \le 1.
\label{eq:gc-soma-control}
\end{equation}
The plus and minus branches generate overshooting and interpolating proposals,
respectively. The implementation varies \(k^\star\) and \(\rho^\star\) with
the consumed evaluation budget and evaluates several nearby contraction
targets during each migration. GC-SOMA therefore retains a discretized
SOMA-like path while expressing its parameters through intended geometric
effects.

\paragraph{Rotation-Aware SOMA.}
Rotation-Aware SOMA (RA-SOMA) addresses the axis alignment of the standard
Bernoulli mask. Let \(Q\) contain the eigenvectors of the current population
covariance matrix. The displacement is transformed into this basis, masked,
and transformed back:
\begin{equation}
z
=
x+
tQ\left[m\cdot Q^{\mathsf T}(L-x)\right].
\label{eq:ra-soma-design}
\end{equation}

Equivalently, the diagonal mask is replaced by the correlated operator
\(Q\operatorname{diag}(m)Q^{\mathsf T}\). The basis is recomputed during the
optimization, so the active proposal directions follow the orientation
represented by the current population rather than the fixed coordinate axes.

Covariance-derived coordinate systems have previously been used to reduce
coordinate dependence in evolutionary search. In DE, eigenvector-based
crossover applies recombination in a basis estimated from the population
covariance \cite{guo2015eigenvector}, and rotationally invariant DE operators
have been studied systematically \cite{caraffini2019rotation}. More broadly,
invariance can materially affect performance on ill-conditioned and
nonseparable problems \cite{hansen2011invariance}. RA-SOMA transfers this
principle specifically to the SOMA perturbation mask: the leader-directed
proposal remains unchanged in form, but coordinate activation is performed in
a population-derived basis.

\paragraph{Experimental iL-SHOMA-RA.}
The third method is an experimental composite optimizer constructed on top of
iSOMA. It retains the rank-based migrant selection, tournament-based leader
selection, decreasing iSOMA path, first-improvement acceptance, and stagnation
restart. The success-history and population-reduction components are motivated by the
SHADE and L-SHADE design principles
\cite{tanabe2013shade,tanabe2014lshade}. Their transfer to a
rotation-aware SOMA mechanism nevertheless produces a new composite
optimizer rather than a variant or implementation of iL-SHADE.

Three mechanisms are added:

\begin{enumerate}
    \item the perturbation mask is applied in the current population covariance
    basis as in RA-SOMA;
    \item the perturbation probability and a multiplicative path scale are
    sampled from success-history memories; and
    \item the population size is reduced linearly during the run.
\end{enumerate}

For memory index \(r\), the sampled parameters have the form
\begin{equation}
p
\sim
\mathcal{N}\!\left(
\alpha M_{p,r}+(1-\alpha)p_{\mathrm{iSOMA}},
\sigma_p^2
\right),
\qquad
s
\sim
\operatorname{Cauchy}(M_{s,r},\gamma_s),
\qquad
t=s\,t_{\mathrm{iSOMA}}.
\label{eq:ilshoma-parameter-sampling}
\end{equation}
Only parameter pairs associated with accepted improvements update the
memories. The perturbation probability uses an improvement-weighted arithmetic
mean, whereas the path scale uses an improvement-weighted Lehmer mean. The
population size follows
\begin{equation}
N(q)
=
\max\left\{
N_{\min},
\operatorname{round}\!\left[
N_0+(N_{\min}-N_0)\frac{q}{B}
\right]
\right\},
\label{eq:ilshoma-population-reduction}
\end{equation}
where \(q\) is the number of consumed evaluations and \(B\) is the evaluation
budget. We refer to this composite method as iL-SHOMA-RA. Because several
mechanisms are introduced simultaneously, it is treated as an exploratory
optimizer rather than as evidence for the effect of any single component.

\paragraph{Reference methods and benchmark design.}
The comparison includes canonical All-to-One SOMA, the supplied iSOMA
implementation, SciPy differential evolution using the
\texttt{best1bin} strategy \cite{virtanen2020scipy}, and the
\texttt{pyade-python} implementation of iL-SHADE
\cite{brest2016ilshade,pyade2019}. GC-SOMA, RA-SOMA,
iSOMA, iL-SHOMA-RA, SciPy-DE, and iL-SHADE use an initial population of
\(4D\). Canonical SOMA uses its conventional population setting of \(10D\).
The matched SOMA variants and SciPy-DE receive a common initial population
within each experimental block. For iL-SHADE, the initial population size is
overridden to $4D$, while population initialization remains internal to the
PyADE implementation.

Following established benchmarking principles
\cite{more2009benchmarking,beiranvand2017bestpractices}, the algorithms were
evaluated on the complete noiseless BBOB suite
\(f_1\)--\(f_{24}\), comprising separable functions, functions with low or
moderate conditioning, highly conditioned unimodal functions, multimodal
functions with adequate global structure, and multimodal functions with weak
global structure \cite{hansen2021coco,finck2009bbob,hansen2009bbobsetup, jamil2013benchmarkfunctions}. Dimensions
\[
D\in\{5,10,20\}
\]
and budgets
\[
B\in\{500D,2500D\}
\]
were used. Each function was evaluated on five BBOB instances with ten
independent optimizer repetitions per instance. Each algorithm therefore
completed \(24\times5\times10=1200\) runs for every dimension--budget
configuration.

The primary performance value is the final absolute error
\begin{equation}
\Delta f=f(x_{\mathrm{best}})-f_{\mathrm{opt}}.
\label{eq:bbob-absolute-error}
\end{equation}
Repeated runs were first aggregated by their median within each
function--instance block. Average ranks were then calculated over the resulting
120 blocks, with lower ranks indicating better performance. The Friedman test assessed overall differences. Pairwise comparisons
used the Wilcoxon signed-rank test with Holm correction
\cite{derrac2011nonparametric,garcia2010nonparametric}.

Anytime performance was measured by the normalized absolute-error area under
the best-so-far curve,
\begin{equation}
A_{\mathrm{abs}}
=
\frac{1}{B}
\sum_{q=1}^{B}
\min\left\{
\frac{\Delta f_q}
     {\max(\Delta f_1,\varepsilon)},
1
\right\}.
\label{eq:absolute-auc}
\end{equation}
Lower values indicate that small errors were obtained earlier and maintained
over a larger proportion of the evaluation budget. Final error remains the
primary endpoint; \(A_{\mathrm{abs}}\), target attainment, and convergence
curves describe the behavior during the run.

\subsection{Numerical results}
\label{subsec:numerical-results}

\paragraph{Overall ranking}
Table~\ref{tab:overall-bbob-ranks} reports the average final-error ranks.
Canonical SOMA was ranked last or nearly last in every configuration, whereas
all operator-guided and improved SOMA variants were substantially more
competitive. The Friedman test rejected the null hypothesis of equal rank distributions in every
dimension--budget configuration, with all overall \(p\)-values below
\(10^{-33}\).

\begin{table*}[t]
\centering
\caption{Average ranks over the 120 BBOB function--instance blocks.
Lower values are better. The best result in each column is shown in bold.}
\label{tab:overall-bbob-ranks}
\small
\setlength{\tabcolsep}{5pt}
\begin{tabular}{lrrrrrr}
\toprule
& \multicolumn{2}{c}{\(D=5\)}
& \multicolumn{2}{c}{\(D=10\)}
& \multicolumn{2}{c}{\(D=20\)} \\
\cmidrule(lr){2-3}
\cmidrule(lr){4-5}
\cmidrule(lr){6-7}
Algorithm
& \(500D\) & \(2500D\)
& \(500D\) & \(2500D\)
& \(500D\) & \(2500D\) \\
\midrule
Canonical-SOMA
& 6.833 & 5.888 & 6.933 & 6.588 & 6.842 & 6.596 \\
GC-SOMA
& 4.067 & 4.112 & 4.467 & 4.362 & 4.537 & 4.521 \\
RA-SOMA
& 3.175 & \textbf{2.596} & 3.892 & 3.233 & 3.929 & 3.996 \\
iSOMA
& 4.408 & 4.204 & 4.088 & 3.917 & 3.325 & 3.533 \\
iL-SHOMA-RA
& 3.663 & 3.725 & 3.246 & 3.983 & 2.675 & 3.667 \\
SciPy-DE
& \textbf{2.683} & 3.650 & 2.900 & 3.117 & 4.771 & 3.379 \\
iL-SHADE
& 3.171 & 3.825 & \textbf{2.475} & \textbf{2.800}
& \textbf{1.921} & \textbf{2.308} \\
\bottomrule
\end{tabular}
\end{table*}

RA-SOMA produced the best overall rank at \(D=5\) and \(2500D\)
evaluations. At \(D=10\), RA-SOMA remained competitive with the DE
algorithms under the larger budget, but iL-SHADE obtained the best rank at
both budgets. At \(D=20\), iL-SHADE was the strongest general method,
although iL-SHOMA-RA ranked second under the restricted \(500D\) budget.

\paragraph{iL-SHOMA-RA versus iSOMA}
Table~\ref{tab:ilshoma-isoma-pairwise} isolates whether the composite method
improves its iSOMA base. Under \(500D\) evaluations, iL-SHOMA-RA
significantly outperformed iSOMA at all three dimensions. The advantage
remained significant at \(D=5\) under \(2500D\), but not at \(D=10\) or
\(D=20\). The combined method therefore improves the limited-budget behavior
of iSOMA more consistently than its longer-budget final precision.

\begin{table}[t]
\centering
\caption{Paired comparison of iL-SHOMA-RA against iSOMA over 120
function--instance blocks. W/T/L gives wins, ties, and losses of
iL-SHOMA-RA. The ratio is the median blockwise final-error ratio; values below
one favor iL-SHOMA-RA.}
\label{tab:ilshoma-isoma-pairwise}
\small
\begin{tabular}{rrrrr}
\toprule
\(D\) & Budget & W/T/L & Error ratio & \(p_{\mathrm{Holm}}\) \\
\midrule
5  & \(500D\)  & 75/13/32 & 0.810 & \(1.35\times10^{-3}\) \\
5  & \(2500D\) & 66/18/36 & 0.889 & \(3.80\times10^{-4}\) \\
10 & \(500D\)  & 79/11/30 & 0.866 & \(4.28\times10^{-4}\) \\
10 & \(2500D\) & 57/16/47 & 1.000 & \(1.00\) \\
20 & \(500D\)  & 77/11/32 & 0.824 & \(5.55\times10^{-4}\) \\
20 & \(2500D\) & 50/11/59 & 1.000 & \(0.858\) \\
\bottomrule
\end{tabular}
\end{table}

The improvement was concentrated on landscapes for which oriented search
directions are expected to be useful. On the highly conditioned unimodal
group under \(500D\), iL-SHOMA-RA obtained W/T/L counts of
\(20/0/5\), \(23/0/2\), and \(25/0/0\) for
\(D=5,10,20\), respectively. The corresponding corrected \(p\)-values were
\(0.0203\), \(0.00263\), and \(1.25\times10^{-6}\). At \(D=20\), its
median blockwise error was approximately \(9.3\%\) of that of iSOMA on this
group. Significant advantages were also observed on the structured multimodal
group at \(D=10\) and \(D=20\) under \(500D\). No similarly consistent
advantage occurred on separable functions or on multimodal functions with weak
global structure.

\paragraph{Low-dimensional performance of RA-SOMA}
The best RA-SOMA configuration was not only highly ranked but also
statistically better than both DE references. At \(D=5\) and \(2500D\),
RA-SOMA won 75 of 120 blocks against iL-SHADE and 72 against SciPy-DE.
Its median blockwise error was approximately \(32\%\) of the iL-SHADE error
and \(61\%\) of the SciPy-DE error.

\begin{table}[t]
\centering
\caption{Pairwise final-error comparisons for RA-SOMA at \(D=5\) and
\(2500D\) evaluations. W/T/L is reported from the perspective of RA-SOMA.}
\label{tab:ra-soma-highlight}
\small
\begin{tabular}{lrrr}
\toprule
Comparator & W/T/L & Error ratio & \(p_{\mathrm{Holm}}\) \\
\midrule
SciPy-DE  & 72/21/27 & 0.605 & \(7.96\times10^{-6}\) \\
iL-SHADE & 75/15/30 & 0.324 & \(6.07\times10^{-3}\) \\
\bottomrule
\end{tabular}
\end{table}

At $D=10$, no statistically significant difference was detected between
RA-SOMA and the DE references under the larger budget, whereas RA-SOMA was
inferior to iL-SHADE at $D=20$. The benefit of the
population covariance basis is therefore substantial but dimension-dependent.

\paragraph{Anytime behavior}
Table~\ref{tab:auc-function-wins} counts the functions on which an algorithm
obtained the lowest median normalized absolute-error AUC. SciPy-DE dominated
this measure at $D=5$, whereas iL-SHOMA-RA became the leading
anytime method as dimension increased. It had the best AUC on 17 of 24
functions at \(D=20\) and \(500D\), and on 11 functions at \(D=20\) and
\(2500D\).

\begin{table}[t]
\centering
\caption{Algorithm with the lowest median normalized absolute-error AUC on
the largest number of BBOB functions in each configuration.}
\label{tab:auc-function-wins}
\small
\begin{tabular}{rrlr}
\toprule
\(D\) & Budget & Algorithm & Function wins \\
\midrule
5  & \(500D\)  & SciPy-DE      & 17/24 \\
5  & \(2500D\) & SciPy-DE      & 14/24 \\
10 & \(500D\)  & iL-SHOMA-RA   & 9/24 \\
10 & \(2500D\) & SciPy-DE      & 12/24 \\
20 & \(500D\)  & iL-SHOMA-RA   & 17/24 \\
20 & \(2500D\) & iL-SHOMA-RA   & 11/24 \\
\bottomrule
\end{tabular}
\end{table}

The low AUC of iL-SHOMA-RA was not explained solely by rapid progress to a
poor plateau. Table~\ref{tab:strict-target-success} shows selected success
rates for the strict target \(\Delta f\leq10^{-8}\). RA-SOMA had the highest
attainment rate in the low-dimensional, high-budget configuration.
iL-SHOMA-RA had the highest rate at \(D=10\) and \(D=20\) under the
restricted budget, exceeding both iSOMA and iL-SHADE.

\begin{table*}[t]
\centering
\caption{Percentage of runs attaining
\(\Delta f\leq10^{-8}\) in selected configurations. The best result in each
column is shown in bold.}
\label{tab:strict-target-success}
\small
\begin{tabular}{lrrr}
\toprule
Algorithm
& \(D=5,\;2500D\)
& \(D=10,\;500D\)
& \(D=20,\;500D\) \\
\midrule
Canonical-SOMA & 10.0 & 2.0 & 0.8 \\
GC-SOMA        & 20.6 & 5.4 & 4.9 \\
RA-SOMA        & \textbf{52.4} & 5.0 & 3.4 \\
iSOMA           & 27.8 & 9.8 & 9.2 \\
iL-SHOMA-RA     & 22.7 & \textbf{13.8} & \textbf{10.1} \\
SciPy-DE        & 42.1 & 6.7 & 3.2 \\
iL-SHADE        & 21.6 & 12.2 & 7.0 \\
\bottomrule
\end{tabular}
\end{table*}

\paragraph{Geometry control}
GC-SOMA consistently improved upon canonical SOMA but did not obtain the
best aggregate rank in any configuration. This result separates
interpretability from optimization dominance: the analytical formulas provide
direct control over expected active dimensionality and contraction, but the
particular budget-dependent schedule used here is not necessarily optimal.
GC-SOMA should therefore be interpreted as a proof of concept for
geometry-based parameterization rather than as a fully tuned state-of-the-art
optimizer.

Complete function-wise convergence profiles are provided in
Appendix~\ref{app:convergence} for every BBOB function, dimension, and
evaluation budget. The profiles retain the individual function scales rather
than pooling absolute errors across heterogeneous objectives and therefore
provide a trajectory-level complement to the average ranks, paired tests,
normalized AUC values, and target-attainment summaries. They also illustrate
that the relative advantages of the algorithms are landscape- and
budget-dependent rather than uniform across the benchmark. The target-attainment and convergence summaries follow the general
anytime-performance perspective used in black-box optimization benchmarking
\cite{hansen2022anytime}. The normalized absolute-error AUC used here is an
additional within-run progress measure and is not intended as a replacement
for the standard target-based COCO indicators.

\section{Discussion}
\label{sec:discussion}

The operator--selection factorization proved useful not only as an analytical
description but also as a design principle. The SOMA analysis exposed two
actionable properties: the indirect relation between algorithmic parameters
and proposal geometry, and the coordinate dependence introduced by
axis-aligned masking. GC-SOMA and RA-SOMA address these properties in different
ways, showing that proposal-level quantities can be converted into explicit
algorithmic controls. Their nonuniform performance also confirms that proposal
geometry constrains the candidates supplied to selection but does not alone
determine the resulting population dynamics.

The performance of the RA-SOMA configuration is consistent with a practical
benefit from population-aligned masking, although the present comparison
does not isolate the basis transformation from the accompanying population,
path, and repair settings. Its population-derived basis
aligns masked movement with directions represented by the current population,
which is particularly plausible on conditioned and nonseparable landscapes.
The weaker behavior observed as dimension increased suggests, however, that a
full empirical covariance basis is not uniformly reliable. With a finite and
progressively selected population, its eigendirections may become noisy or
overly restrictive. More robust alternatives include shrinkage estimates,
low-rank subspaces, mixtures of coordinate and eigenvector masking, and less
frequent or incremental basis updates. Related work on efficient
eigenvector-based crossover indicates that rank-one updates may reduce the cost
of repeatedly reconstructing such bases
\cite{choi2025eigenvector}.

GC-SOMA illustrates a separate distinction between interpretability and
adaptation quality. Expected active dimensionality and leader contraction are
meaningful control variables, but specifying them geometrically does not by
itself determine an effective schedule. Their values may need to respond to
selection success, population diversity, stagnation, and the remaining
evaluation budget. GC-SOMA should therefore be regarded as evidence that the
derived quantities are operationally usable, rather than as evidence that the
particular schedule examined here is optimal.

The composite iL-SHOMA-RA method appears primarily to improve evaluation
efficiency rather than uniformly improve eventual precision. Its combination
of rotation-aware masking, success-history adaptation, and population-size
reduction can concentrate evaluations on productive proposals, but the present
design does not identify which component is responsible for the observed
behavior. Component-level conclusions therefore require an ablation comparing
the complete method with variants that introduce each mechanism separately and
in selected combinations. Such experiments would also clarify whether the
mechanisms are complementary or whether one accounts for most of the gain.

The DE methods remain an important reference because they show that
analytically motivated SOMA modifications do not imply general dominance over
established adaptive optimizers. More broadly, the results emphasize that raw
proposal geometry interacts with target construction, population sampling,
boundary repair, acceptance rules, adaptation, and population management. The
current evidence is also limited to noiseless, unconstrained BBOB functions
and dimensions up to \(20\). Blockwise ranks, paired tests, target attainment,
and convergence profiles should remain the primary evidence because absolute
errors are not directly comparable across heterogeneous functions, while the
normalized absolute-error AUC is only a within-run progress measure.
Computational overhead may also matter when objective evaluations are
inexpensive. Future work should therefore examine larger-dimensional,
constrained, and noisy problems, including variational quantum landscapes in
which sampling noise can alter apparent minima and optimizer rankings
\cite{illesova2025vhaoptimization,
novak2025reliablevqa,novak2026vqaoptimizers}. These extensions should determine when proposal-level
geometry is sufficient and when it must be coupled to stronger adaptation,
selection, and population-management mechanisms.

\section{Conclusion}
\label{sec:conclusion}

This paper introduced an operator--selection factorization for the analysis of
population-based optimization algorithms. The central observation is that the
nonlinearity of a complete optimization iteration does not preclude exact
linear or affine structure within its variation stage. Candidate generation
can be isolated as an objective-independent proposal operator once the
required population state and random choices have been fixed, while boundary
repair, leader selection, survivor replacement, ranking, and adaptation remain
separate nonlinear transformations.

For SOMA, the canonical path proposal was represented exactly as a linear
transformation of an augmented migrant--leader state. In leader-relative
coordinates, the operator reduces to coordinate-wise rescaling and exposes the
geometric roles of interpolation, projection, overshooting, and masking.
Treating the perturbation mask as a Bernoulli random operator yielded
closed-form expressions for the proposal mean, covariance, expected step
length, expected distance from the leader, active dimensionality, and
coordinate coverage. The analysis also clarified how conditioning the mask on
nonzero activation introduces cross-coordinate dependence. For
\texttt{DE/rand/1/bin}, the same viewpoint produced finite-population moments
for differential mutation and separated mutation-induced covariance from the
additional uncertainty and dependence introduced by forced-coordinate
binomial crossover.

These analytical quantities were then used as design variables rather than
only descriptive measures. Geometry-Controlled SOMA parameterized proposals
through intended active dimensionality and leader contraction, while
Rotation-Aware SOMA applied perturbation masks in a population-derived
covariance basis. An experimental extension of iSOMA additionally combined
rotation-aware masking with success-history adaptation and linear
population-size reduction. On the complete noiseless BBOB benchmark, all
operator-guided variants substantially improved upon canonical SOMA.
Rotation-Aware SOMA was particularly effective in the low-dimensional,
high-budget regime, whereas the composite iSOMA extension showed its clearest
advantages in restricted-budget and anytime performance. The improvements
were nevertheless dependent on dimension, function group, and evaluation
budget, and the proposed methods did not uniformly dominate the DE reference
algorithms.

The results therefore support proposal-level geometry as a useful connection
between algorithm analysis and algorithm design, while also showing that
proposal structure alone does not determine complete optimizer behavior.
Selection, target construction, boundary handling, adaptation, and population
management determine how the available geometry is exploited. Future work
should examine component-level ablations of the composite method, more robust
low-rank or regularized rotation models, adaptive geometry-control schedules,
and extensions to larger-dimensional, noisy, and constrained optimization
problems.

\section*{Acknowledgements}
This project has received funding from the Research Council of Lithuania (LMTLT), agreement No. P-ITP-24-9. This research was also supported by research grants SGS No. SP2026/063 of VSB-Technical University of Ostrava, Czech Republic.

\section*{Data availability}
The source code, numerical-verification experiments, BBOB benchmark workflow,
processed results, and figure-generation scripts are available at \url{https://github.com/VojtechNovak/SOMA-operator}. Further details are available through the corresponding author.

\appendix

\section{Numerical Verification and Practical Distortions}
\label{sec:numerical}

The analytical results were evaluated using exact enumeration whenever the
finite support was computationally manageable and Monte Carlo simulation
otherwise. All pseudorandom experiments used the fixed seed
\(20260319\). Each Monte Carlo moment estimate used \(150\,000\)
samples, the coordinate-coverage study used \(60\,000\) independent
trials, and the boundary and selection studies used \(150\,000\) and
\(50\,000\) samples, respectively. Covariance matrices were calculated
using population normalization \(1/n\), matching the probabilistic
definition used in the analytical derivations.

For SOMA, the independent-mask experiments covered dimensions
\(D\in\{3,10,30\}\), activation probabilities
\(p\in\{0.1,0.3,0.5,0.8\}\), and path parameters
\(t\in\{0.5,1,1.5,2.5\}\), with five independently generated
migrant--leader pairs for each parameter combination. The conditioned
mask study used \(D\in\{3,10,30\}\) and
\(p\in\{0.05,0.2,0.5\}\). For DE, ordered and distinct donor indices
were sampled uniformly without replacement from the target-excluded
population. Standard binomial crossover was implemented with one uniformly
selected forced coordinate \(j_{\mathrm{rand}}\).

The experiments were designed to verify distributional identities rather
than to compare optimizer rankings. Boundary repair and fitness-based
selection were then examined separately to quantify how these operations
distort the raw proposal geometry.

\subsection{Verification of SOMA proposal moments}

Under independent Bernoulli masks, the empirical proposal moments closely
matched the analytical expressions
\[
    \mathbb{E}[\bm z]
    =
    \bm x+pt(\bm L-\bm x)
\]
and
\[
    \operatorname{Cov}(\bm z)
    =
    p(1-p)t^2
    \operatorname{diag}
    \left((\bm L-\bm x)^{2}\right).
\]
Across \(240\) primary parameter cases, the mean relative error of the
empirical proposal mean was \(1.20\times10^{-3}\), and the mean relative
Frobenius error of the covariance was \(6.50\times10^{-3}\). The
corresponding maximum errors were \(5.82\times10^{-3}\) and
\(1.85\times10^{-2}\). The expected squared step length and expected
squared leader distance agreed with their analytical values with mean
relative errors of \(1.81\times10^{-3}\) and
\(7.09\times10^{-4}\), respectively.

A separate convergence experiment increased the sample count from
\(10^3\) to \(1.5\times10^5\). Over this range, the mean error decreased
from \(2.14\times10^{-2}\) to \(1.01\times10^{-3}\), while the covariance
error decreased from \(4.63\times10^{-2}\) to
\(5.76\times10^{-3}\). These results are consistent with ordinary Monte
Carlo convergence and provide no indication of a systematic discrepancy
between the procedural and analytical proposal distributions.

\subsection{Conditioning SOMA masks on nonzero activation}

When all-zero masks are rejected, the mask distribution is no longer a
product of independent Bernoulli variables. Writing
\[
    A=1-(1-p)^D,
    \qquad
    \alpha=\frac{p}{A},
\]
the conditional mask mean is
\(\mathbb{E}[\pi_d\mid K>0]=\alpha\), and distinct coordinates satisfy
\[
    \operatorname{Cov}(\pi_d,\pi_e\mid K>0)
    =
    -\frac{p^2(1-p)^D}{A^2},
    \qquad d\neq e.
\]
Consequently, for
\(\Gamma=\operatorname{Cov}(\bm\pi\mid K>0)\),
\[
    \mathbb{E}[\bm z\mid K>0]
    =
    \bm x+t\alpha(\bm L-\bm x),
\]
and
\[
    \operatorname{Cov}(\bm z\mid K>0)
    =
    t^2
    \operatorname{Diag}(\bm L-\bm x)
    \Gamma
    \operatorname{Diag}(\bm L-\bm x).
\]

Across nine conditioned-mask cases, the empirical activation probabilities,
means, and covariances agreed with these formulas. The mean relative error
of the proposal mean was \(6.60\times10^{-4}\), and the covariance error
was \(6.23\times10^{-3}\), with a maximum of
\(1.17\times10^{-2}\). The induced dependence is substantial in low
dimension and at small \(p\). For example, at \(D=3\) and \(p=0.05\),
conditioning increases the marginal activation probability from \(0.05\)
to \(0.35057\), while the theoretical off-diagonal mask covariance is
\(-0.10537\); the empirical value was \(-0.10526\). By contrast, at
\(D=30\) and \(p=0.5\), rejection of the all-zero mask has a negligible
effect because \((1-p)^D\) is extremely small.

The coordinate-coverage simulations also confirmed the path-level
expressions. With independently regenerated masks, the probability of
covering a coordinate at least once was
\(1-(1-p)^J\), whereas reusing one mask left the coverage probability
approximately equal to \(p\), independently of \(J\). For example, with
\(D=30\) and \(p=0.2\), the probability that all coordinates were covered
increased from essentially zero at \(J=5\) to \(0.706\) at \(J=20\)
and \(0.9996\) at \(J=50\). The corresponding theoretical values were
\(6.72\times10^{-6}\), \(0.7062\), and \(0.99957\).

\subsection{Finite-population DE mutation}

For a target-excluded donor pool of size \(M=N-1\), let
\(\bm\mu_{-i}\) and \(\bm S_{-i}\) denote the donor-pool mean and
population covariance, with normalization \(1/M\). Uniform sampling of
three ordered, distinct donors gives
\[
    \mathbb{E}[\bm v]
    =
    \bm\mu_{-i},
\]
and
\[
    \operatorname{Cov}(\bm v)
    =
    \left(
        1+2F^2\frac{M}{M-1}
    \right)
    \bm S_{-i}.
\]
The formula was tested for
\(N\in\{6,10,20,50\}\), \(D\in\{3,10\}\), and
\(F\in\{0.2,0.5,0.8,1.0\}\), using two target indices whenever
applicable. Exact enumeration over all ordered donor triples matched the
analytical moments to floating-point precision: the mean scaled error was
\(2.48\times10^{-16}\), and the covariance relative error was
\(1.71\times10^{-15}\). Monte Carlo estimates produced mean scaled and
covariance errors of \(2.30\times10^{-3}\) and
\(4.37\times10^{-3}\), respectively.

The finite-population factor is visible even when the donor-pool covariance
is fixed. For \(N=6\), corresponding to \(M=5\), the covariance multiplier
increases from \(1.10\) at \(F=0.2\) to \(3.50\) at \(F=1\). As
\(N\) grows, the factor approaches \(1+2F^2\), recovering the
large-population limit while retaining the finite-sampling correction.

\subsection{Exact forced-coordinate DE crossover}

The forced coordinate in standard \texttt{DE/rand/1/bin} changes both the
marginal activation probability and the dependence structure of the
crossover mask. For dimension \(D\) and crossover rate
\(q=\mathrm{CR}\), define
\[
    q_\star=q+\frac{1-q}{D}.
\]
After marginalizing over \(j_{\mathrm{rand}}\), the mask covariance
matrix \(\Gamma\) has entries
\[
    \Gamma_{dd}=q_\star(1-q_\star),
    \qquad
    \Gamma_{de}
    =
    -\frac{(1-q)^2}{D^2},
    \quad d\neq e.
\]
For a fixed target and mutant with
\(\bm\delta=\bm v-\bm x_i\), the exact conditional moments are
\[
    \mathbb{E}[\bm u_i\mid\bm v]
    =
    \bm x_i+q_\star\bm\delta
\]
and
\[
    \operatorname{Cov}(\bm u_i\mid\bm v)
    =
    \operatorname{Diag}(\bm\delta)
    \Gamma
    \operatorname{Diag}(\bm\delta).
\]

Exact enumeration was performed for
\(D\in\{2,3,5,10\}\) and
\(\mathrm{CR}\in\{0,0.1,0.5,0.9,1\}\). The exact mean and covariance
errors were at floating-point precision, while Monte Carlo mean and
covariance errors averaged \(7.91\times10^{-4}\) and
\(3.64\times10^{-3}\). The discrepancy from the independent-mask
approximation is largest at small dimension and low crossover rate. For
example, at \(D=3\) and \(\mathrm{CR}=0.1\), the effective activation
probability is \(q_\star=0.4\), and the off-diagonal mask covariance is
\(-0.09\). Thus, treating the mask entries as independent Bernoulli
variables with activation probability \(0.1\) substantially
mischaracterizes the standard crossover distribution.

\subsection{Combined DE trial moments and parameter effects}

Let
\[
    \bm\delta
    =
    \mathbb{E}[\bm v]-\bm x_i.
\]
Combining donor randomness with forced crossover yields
\[
    \mathbb{E}[\bm u_i]
    =
    \bm x_i+q_\star\bm\delta
\]
and
\[
    \operatorname{Cov}(\bm u_i)
    =
    q_\star^2\operatorname{Cov}(\bm v)
    +
    \Gamma\odot
    \left[
        \operatorname{Cov}(\bm v)
        +
        \bm\delta\bm\delta^\top,
    \right]
\]
where $\odot$ denotes the Hadamard product. The combined expression was tested in \(27\) cases covering three population--dimension pairs,
\(F\in\{0.3,0.6,0.9\}\), and
\(\mathrm{CR}\in\{0.2,0.5,0.9\}\). Exact enumeration again matched the
closed form to numerical precision, with covariance error below
\(9.01\times10^{-15}\). The Monte Carlo trial-mean scaled error averaged
\(2.20\times10^{-3}\), and the covariance error averaged
\(6.45\times10^{-3}\), with a maximum of \(1.13\times10^{-2}\).

The parameter study confirms that \(F\) primarily controls mutation
spread, whereas \(\mathrm{CR}\) controls how strongly that geometry is
transferred to the trial distribution. In the \(N=30,D=10\) population,
at \(\mathrm{CR}=0.5\), increasing \(F\) from \(0.3\) to \(0.9\)
increased the trial covariance trace from \(9.22\) to \(19.55\). At
\(F=0.9\), increasing \(\mathrm{CR}\) from \(0.2\) to \(0.9\)
increased the trace from \(10.25\) to \(31.14\). The principal-axis
angle between mutant and trial covariance decreased from \(23.82^\circ\)
to \(1.91^\circ\) over the same crossover-rate change, showing that
high crossover rates make the trial geometry closely inherit the mutant
orientation.

The closed-form proposal moments describe raw proposals before feasibility
repair. To quantify the resulting limitation, raw SOMA and DE proposals
were generated near the boundary of
\(\Omega=[-1,1]^D\) and repaired by clipping or repeated reflection.
Both operators changed the proposal means and covariances, and the effect
increased rapidly with the proportion of infeasible samples.

For SOMA with \(D=10\), \(p=0.8\), and \(t=3\), all sampled raw
proposals required repair. Clipping retained only \(27.1\%\) of the raw
covariance trace, while reflection retained \(8.7\%\). For DE with
\(D=10\), \(F=1\), and \(\mathrm{CR}=0.9\), \(95.6\%\) of trials
required repair; clipping and reflection retained \(63.0\%\) and
\(44.8\%\) of the raw covariance trace, respectively. At less aggressive
settings, the distortion was considerably smaller. These results confirm
that repair cannot be treated as a negligible implementation detail in
high-spread regimes.

Finally, raw proposal distributions were compared with the distributions
remaining after standard fitness-based selection on the sphere function
and on a rotated quadratic objective. For DE, greedy target--trial
selection produced acceptance rates between \(0.437\) and \(0.962\).
The selected covariance trace ranged from \(42.3\%\) to \(96.0\%\) of
the raw trial trace, depending on dimension, objective, \(F\), and
\(\mathrm{CR}\). The higher-spread setting
\((F,\mathrm{CR})=(0.9,0.9)\) generally produced a larger discrepancy
between pre-selection and post-selection covariance.

The SOMA experiment selected the best point from a discretized
leader-directed proposal family. Because the controlled leader was placed
closer to the minimizer, at least one proposal was selected in almost
every trial. Even in this favorable setting, selection substantially
reshaped the distribution: the selected-to-raw covariance-trace ratio
ranged from \(0.081\) to \(1.410\). Selection therefore need not act as
a uniform contraction of the proposal cloud; it can suppress, preserve,
or amplify different directions depending on the objective and candidate
family.

\subsection{Summary}

The exact-enumeration results verify the finite-population DE donor and
forced-crossover formulas to floating-point precision, while the Monte
Carlo studies validate the SOMA and DE moment expressions with covariance
errors generally below \(2\%\). More importantly, the conditioned-mask,
boundary, and selection experiments identify where the raw linear-operator
description must be qualified. Rejecting all-zero masks and forcing a DE
crossover coordinate introduce cross-coordinate dependence, while repair
and selection can substantially alter both proposal means and covariance.
The operator formulas therefore provide an exact description of the raw
variation stage under their stated assumptions, but not of the complete
feasible or selected update.

\section{Function-wise convergence profiles}
\label{app:convergence}

This appendix reports function-wise convergence profiles for all 24 noiseless
BBOB functions, organized by the five standard function groups and the two
evaluation budgets. Each figure combines the results for
$D\in\{5,10,20\}$ for one function group and budget. The curves show the
median best-so-far absolute error over all instances and independent
repetitions, while the shaded regions indicate the corresponding
interquartile ranges. These profiles complement the aggregate ranks,
statistical comparisons, normalized AUC values, and target-attainment results
reported in Section~\ref{subsec:numerical-results} by showing when performance differences
emerge and whether they persist on individual functions.

% Budget 500D
\bbobconvergencefigure{1}{Separable functions}{group1_separable}{500}
\bbobconvergencefigure{2}{Low or moderate conditioning}{group2_low_moderate_conditioning}{500}
\bbobconvergencefigure{3}{High conditioning and unimodal}{group3_high_conditioning_unimodal}{500}
\bbobconvergencefigure{4}{Multimodal with adequate global structure}{group4_multimodal_adequate_structure}{500}
\bbobconvergencefigure{5}{Multimodal with weak global structure}{group5_multimodal_weak_structure}{500}

% Budget 2500D
\bbobconvergencefigure{1}{Separable functions}{group1_separable}{2500}
\bbobconvergencefigure{2}{Low or moderate conditioning}{group2_low_moderate_conditioning}{2500}
\bbobconvergencefigure{3}{High conditioning and unimodal}{group3_high_conditioning_unimodal}{2500}
\bbobconvergencefigure{4}{Multimodal with adequate global structure}{group4_multimodal_adequate_structure}{2500}
\bbobconvergencefigure{5}{Multimodal with weak global structure}{group5_multimodal_weak_structure}{2500}

\section{Implementation and parameter settings}
\label{app:parameters}

The experiments used the noiseless COCO/BBOB functions
$f_1$--$f_{24}$, dimensions $D\in\{5,10,20\}$, instances $1$--$5$, and
budgets $B\in\{500D,2500D\}$. Ten independent repetitions were performed per
function--instance block. All pseudorandom seeds were derived deterministically
from the base seed \(20260723\), the dimension, budget, function, instance, and
repetition using NumPy \texttt{SeedSequence}. Initial populations were sampled
uniformly within the COCO bounds. GC-SOMA, RA-SOMA, iSOMA,
iL-SHOMA-RA, and SciPy-DE used population size \(4D\); the first four and
SciPy-DE received the same initial population within each block, while
iL-SHADE used its internally generated seeded population. Canonical SOMA used
a separately seeded population of size \(10D\). Initial-population evaluations
were included in the budget, and a common objective wrapper enforced an exact
maximum of \(B\) true function evaluations.

\begin{table}[p]
\centering
\footnotesize
\caption{Optimizer configurations used in the BBOB experiments. Masks were
regenerated independently at every proposal point and were permitted to be
all zero.}
\label{tab:optimizer-parameters}
\renewcommand{\arraystretch}{1.12}
\setlength{\tabcolsep}{4pt}
\begin{tabularx}{\textwidth}{@{}lX@{}}
\toprule
Method & Configuration \\
\midrule

Canonical-SOMA &
All-to-One with a fixed leader during each migration and a common origin for
all path points. Population \(10D\), \(\mathrm{PRT}=0.10\),
\(\mathrm{Step}=0.11\), and \(\mathrm{PathLength}=3.0\), giving
\(t\in\{0.11,0.22,\ldots,2.97\}\). Each migrant retained its best path point
only when it improved upon the origin. Boundary violations were clipped. \\

GC-SOMA &
Population \(4D\), identity masking basis, six proposals per migrant, and
reflection repair. For consumed-budget fraction \(s=q/B\),
\[
p=0.35+0.55s,\qquad
\rho_c=\max\{0.85-0.70s,\,1-p+10^{-6}\}.
\]
For
\(\delta_j\in\operatorname{linspace}(0.12,-0.12,6)\),
\[
\rho_j=\operatorname{clip}
 \bigl(\rho_c+\delta_j,\,
       \max\{1-p+10^{-6},0.02\},\,0.98\bigr),
\]
and
\[
t_j=1\pm\sqrt{1-\frac{1-\rho_j}{p}}.
\]
The overshooting branch was used for \(s<0.55\), and the interpolating branch
thereafter. The best improving proposal was retained. \\

RA-SOMA &
Population \(4D\), \(\mathrm{PRT}=0.30\), and six equally spaced path values
\(t\in\{0.4,0.8,1.2,1.6,2.0,2.4\}\). At the start of each migration, the
eigenvectors of the current population covariance formed the masking basis.
The covariance used normalization \(1/N\) and regularization
\(10^{-12}(\operatorname{tr}\Sigma/D)I\), or \(10^{-12}I\) when the trace
vanished. Reflection repair and best-path-point acceptance were used. \\

iSOMA &
The supplied \texttt{ISOMAOptimizer} implementation was loaded without
modification. Population \(4D\),
\(N_{\mathrm{jump}}=10\), \(\mathrm{Step}=0.3\),
\(\mathrm{MaxMigration}=10000\), \(m=10\), \(n=5\), and \(k=15\).
The maximum function evaluations and COCO bounds were passed directly to the
implementation. Its seeded \texttt{numpy.random.rand} initialization was
matched to the common \(4D\) population. \\

iL-SHOMA-RA &
Initial population \(N_0=4D\), \(N_{\min}=\max\{4,D\}=D\), with
\(N_{\mathrm{jump}}=10\), \(\mathrm{Step}=0.3\), and initial selection
parameters \(m=10\), \(n=5\), \(k=15\); their proportions were preserved as
the population decreased. The covariance basis was recomputed each migration.
Six-element memories were initialized to \(M_p=0.5\) and \(M_s=1\).
For a randomly selected memory entry,
\[
p=\operatorname{clip}\!\left(
 \mathcal{N}\!\left(
  0.7M_p+0.3[0.1+0.9q/B],\,0.1^2
 \right),\,1/D,\,1
\right),
\]
while the path scale was sampled from
\(\operatorname{Cauchy}(M_s,0.1)\), restricted to \(0.05<s\leq2\).
For jump \(j=1,\ldots,10\),
\(t=s(10-j+1)0.3\). First improvement was accepted. Successful \(p\) values
updated the memory by an improvement-weighted arithmetic mean and successful
scales by an improvement-weighted Lehmer mean. The population was reduced
linearly while retaining the best individuals. After more than \(50N\)
unsuccessful migrant attempts without improvement, \(10\%\) of the
non-best population was resampled uniformly. Boundary violations were clipped. \\

SciPy-DE &
SciPy \texttt{differential\_evolution} with population \(4D\) supplied through
\texttt{init}, strategy \texttt{best1bin}, differential weight \(F=0.8\),
crossover rate \(\mathrm{CR}=0.9\), immediate updating, one worker, no
polishing, and \(\mathrm{tol}=\mathrm{atol}=0\). The SciPy population
multiplier was \(4\), and the iteration limit was
\(\lceil(B-4D)/(4D)\rceil\); the common objective wrapper enforced the exact
budget. \\

iL-SHADE &
PyADE \texttt{ilshade.get\_default\_params(D)} was used, with only the
objective, COCO bounds, random seed, budget \(B\), and initial population size
\(4D\) overridden. Population initialization remained internal to PyADE.
Callbacks and optional arguments were disabled, and no fallback
implementation was used. \\

\bottomrule
\end{tabularx}
\end{table}

The reported objective value was the absolute BBOB error
\(\Delta f=\max\{f(x)-f_{\mathrm{opt}},0\}\). Target attainment was evaluated
at
\[
10^{2},10^{1},10^{0},10^{-1},\ldots,10^{-8}.
\]
Best-so-far histories were retained at every true evaluation; the published
convergence files used the union of 201 linearly spaced and 201 geometrically
spaced evaluation points. The numerical floor used for logarithms and the
normalized AUC was \(10^{-300}\).

\scriptsize
\bibliographystyle{elsarticle-num}
\bibliography{bibfile,bibfile_revisions}

\begin{thebibliography}{10}
\expandafter\ifx\csname url\endcsname\relax
  \def\url#1{\texttt{#1}}\fi
\expandafter\ifx\csname urlprefix\endcsname\relax\def\urlprefix{URL }\fi
\expandafter\ifx\csname href\endcsname\relax
  \def\href#1#2{#2} \def\path#1{#1}\fi

\bibitem{storn1997differential}
R.~Storn, K.~Price, Differential evolution---a simple and efficient heuristic for global optimization over continuous spaces, Journal of Global Optimization 11~(4) (1997) 341--359.
\newblock \href {https://doi.org/10.1023/A:1008202821328} {\path{doi:10.1023/A:1008202821328}}.

\bibitem{davendra2016soma}
D.~Davendra, I.~Zelinka, Self-Organizing Migrating Algorithm: Methodology and Implementation, Springer International Publishing, Cham, 2016.
\newblock \href {https://doi.org/10.1007/978-3-319-28161-2} {\path{doi:10.1007/978-3-319-28161-2}}.

\bibitem{skanderova2023soma}
L.~Skanderova, Self-organizing migrating algorithm: Review, improvements and comparison, Artificial Intelligence Review 56~(1) (2023) 101--172.
\newblock \href {https://doi.org/10.1007/s10462-022-10167-8} {\path{doi:10.1007/s10462-022-10167-8}}.

\bibitem{eiben2015introduction}
A.~E. Eiben, J.~E. Smith, Introduction to Evolutionary Computing, 2nd Edition, Springer, Berlin, Heidelberg, 2015.
\newblock \href {https://doi.org/10.1007/978-3-662-44874-8} {\path{doi:10.1007/978-3-662-44874-8}}.

\bibitem{zelinka2015survey}
I.~Zelinka, A survey on evolutionary algorithms dynamics and its complexity---mutual relations, past, present and future, Swarm and Evolutionary Computation 25 (2015) 2--14.
\newblock \href {https://doi.org/10.1016/j.swevo.2015.06.002} {\path{doi:10.1016/j.swevo.2015.06.002}}.

\bibitem{doerr2021survey}
B.~Doerr, F.~Neumann, A survey on recent progress in the theory of evolutionary algorithms for discrete optimization, ACM Transactions on Evolutionary Learning and Optimization 1~(4) (2021) 1--43.
\newblock \href {https://doi.org/10.1145/3472304} {\path{doi:10.1145/3472304}}.

\bibitem{das2011survey}
S.~Das, P.~N. Suganthan, Differential evolution: A survey of the state-of-the-art, IEEE Transactions on Evolutionary Computation 15~(1) (2011) 4--31.
\newblock \href {https://doi.org/10.1109/TEVC.2010.2059031} {\path{doi:10.1109/TEVC.2010.2059031}}.

\bibitem{das2016recent}
S.~Das, S.~S. Mullick, P.~N. Suganthan, Recent advances in differential evolution---an updated survey, Swarm and Evolutionary Computation 27 (2016) 1--30.
\newblock \href {https://doi.org/10.1016/j.swevo.2016.01.004} {\path{doi:10.1016/j.swevo.2016.01.004}}.

\bibitem{opara2019survey}
K.~R. Opara, J.~Arabas, Differential evolution: A survey of theoretical analyses, Swarm and Evolutionary Computation 44 (2019) 546--558.
\newblock \href {https://doi.org/10.1016/j.swevo.2018.06.010} {\path{doi:10.1016/j.swevo.2018.06.010}}.

\bibitem{novak2026cec}
V.~Nov{\'a}k, T.~Bezd{\v{e}}k, I.~Zelinka, S.~Das, M.~Beseda, A longitudinal analysis of the {CEC} single-objective competitions (2010--2024) and implications for variational quantum optimization, Swarm and Evolutionary Computation (2026) 102469\href {https://doi.org/10.1016/j.swevo.2026.102469} {\path{doi:10.1016/j.swevo.2026.102469}}.

\bibitem{opara2018mutation}
K.~R. Opara, J.~Arabas, Comparison of mutation strategies in differential evolution---a probabilistic perspective, Swarm and Evolutionary Computation 39 (2018) 53--69.
\newblock \href {https://doi.org/10.1016/j.swevo.2017.12.007} {\path{doi:10.1016/j.swevo.2017.12.007}}.

\bibitem{opara2019contour}
K.~R. Opara, J.~Arabas, The contour fitting property of differential mutation, Swarm and Evolutionary Computation 50 (2019) 100441.
\newblock \href {https://doi.org/10.1016/j.swevo.2018.09.001} {\path{doi:10.1016/j.swevo.2018.09.001}}.

\bibitem{zaharie2009crossover}
D.~Zaharie, Influence of crossover on the behavior of differential evolution algorithms, Applied Soft Computing 9~(3) (2009) 1126--1138.
\newblock \href {https://doi.org/10.1016/j.asoc.2009.02.012} {\path{doi:10.1016/j.asoc.2009.02.012}}.

\bibitem{pluhacek2021explaining}
M.~Pluh{\'a}{\v c}ek, A.~Kazikova, T.~Kadavy, A.~Viktorin, R.~Senkerik, Explaining {SOMA}: The relation of stochastic perturbation to population diversity and parameter space coverage, in: Proceedings of the Genetic and Evolutionary Computation Conference Companion, Association for Computing Machinery, New York, NY, USA, 2021, pp. 1944--1952.
\newblock \href {https://doi.org/10.1145/3449726.3463211} {\path{doi:10.1145/3449726.3463211}}.

\bibitem{biedrzycki2019bounds}
R.~Biedrzycki, J.~Arabas, D.~Jagodzi{\'n}ski, Bound constraints handling in differential evolution: An experimental study, Swarm and Evolutionary Computation 50 (2019) 100453.
\newblock \href {https://doi.org/10.1016/j.swevo.2018.10.004} {\path{doi:10.1016/j.swevo.2018.10.004}}.

\bibitem{tomczak2020reversible}
J.~M. Tomczak, E.~W{\k e}glarz-Tomczak, A.~E. Eiben, Differential evolution with reversible linear transformations, in: Proceedings of the 2020 Genetic and Evolutionary Computation Conference Companion, Association for Computing Machinery, New York, NY, USA, 2020, pp. 205--206, extended version available as arXiv:2002.02869.
\newblock \href {https://doi.org/10.1145/3377929.3389972} {\path{doi:10.1145/3377929.3389972}}.

\bibitem{diep2022isoma}
Q.~B. Diep, T.~C. Truong, S.~Das, I.~Zelinka, Self-organizing migrating algorithm with narrowing search space strategy for robot path planning, Applied Soft Computing 116 (2022) 108270.
\newblock \href {https://doi.org/10.1016/j.asoc.2021.108270} {\path{doi:10.1016/j.asoc.2021.108270}}.

\bibitem{brest2006jde}
J.~Brest, S.~Greiner, B.~Bo{\v{s}}kovi{\'c}, M.~Mernik, V.~{\v{Z}}umer, Self-adapting control parameters in differential evolution: A comparative study on numerical benchmark problems, IEEE Transactions on Evolutionary Computation 10~(6) (2006) 646--657.
\newblock \href {https://doi.org/10.1109/TEVC.2006.872133} {\path{doi:10.1109/TEVC.2006.872133}}.

\bibitem{qin2009sade}
A.~K. Qin, V.~L. Huang, P.~N. Suganthan, Differential evolution algorithm with strategy adaptation for global numerical optimization, IEEE Transactions on Evolutionary Computation 13~(2) (2009) 398--417.
\newblock \href {https://doi.org/10.1109/TEVC.2008.927706} {\path{doi:10.1109/TEVC.2008.927706}}.

\bibitem{zhang2009jade}
J.~Zhang, A.~C. Sanderson, {JADE}: Adaptive differential evolution with optional external archive, IEEE Transactions on Evolutionary Computation 13~(5) (2009) 945--958.
\newblock \href {https://doi.org/10.1109/TEVC.2009.2014613} {\path{doi:10.1109/TEVC.2009.2014613}}.

\bibitem{tanabe2013shade}
R.~Tanabe, A.~S. Fukunaga, Success-history based parameter adaptation for differential evolution, in: 2013 IEEE Congress on Evolutionary Computation, IEEE, 2013, pp. 71--78.
\newblock \href {https://doi.org/10.1109/CEC.2013.6557555} {\path{doi:10.1109/CEC.2013.6557555}}.

\bibitem{tanabe2014lshade}
R.~Tanabe, A.~S. Fukunaga, Improving the search performance of {SHADE} using linear population size reduction, in: 2014 IEEE Congress on Evolutionary Computation, IEEE, 2014, pp. 1658--1665.
\newblock \href {https://doi.org/10.1109/CEC.2014.6900380} {\path{doi:10.1109/CEC.2014.6900380}}.

\bibitem{brest2016ilshade}
J.~Brest, M.~Sepesy~Mau{\v{c}}ec, B.~Bo{\v{s}}kovi{\'c}, {iL-SHADE}: Improved {L-SHADE} algorithm for single-objective real-parameter optimization, in: 2016 IEEE Congress on Evolutionary Computation, IEEE, 2016, pp. 1188--1195.
\newblock \href {https://doi.org/10.1109/CEC.2016.7743922} {\path{doi:10.1109/CEC.2016.7743922}}.

\bibitem{guo2015eigenvector}
S.-M. Guo, C.-C. Yang, Enhancing differential evolution utilizing eigenvector-based crossover operator, IEEE Transactions on Evolutionary Computation 19~(1) (2015) 31--49.
\newblock \href {https://doi.org/10.1109/TEVC.2013.2297160} {\path{doi:10.1109/TEVC.2013.2297160}}.

\bibitem{caraffini2019rotation}
F.~Caraffini, F.~Neri, A study on rotation invariance in differential evolution, Swarm and Evolutionary Computation 50 (2019) 100436.
\newblock \href {https://doi.org/10.1016/j.swevo.2018.08.013} {\path{doi:10.1016/j.swevo.2018.08.013}}.

\bibitem{huxley1942evolution}
J.~Huxley, Evolution: The Modern Synthesis, George Allen \& Unwin, London, 1942.

\bibitem{hansen2001cmaes}
N.~Hansen, A.~Ostermeier, Completely derandomized self-adaptation in evolution strategies, Evolutionary Computation 9~(2) (2001) 159--195.
\newblock \href {https://doi.org/10.1162/106365601750190398} {\path{doi:10.1162/106365601750190398}}.

\bibitem{hansen2011invariance}
N.~Hansen, R.~Ros, N.~Mauny, M.~Schoenauer, A.~Auger, Impacts of invariance in search: When {CMA-ES} and {PSO} face ill-conditioned and non-separable problems, Applied Soft Computing 11~(8) (2011) 5755--5769.
\newblock \href {https://doi.org/10.1016/j.asoc.2011.03.001} {\path{doi:10.1016/j.asoc.2011.03.001}}.

\bibitem{virtanen2020scipy}
P.~Virtanen, R.~Gommers, T.~E. Oliphant, M.~Haberland, T.~Reddy, D.~Cournapeau, E.~Burovski, P.~Peterson, W.~Weckesser, J.~Bright, S.~J. {van der Walt}, M.~Brett, J.~Wilson, K.~J. Millman, P.~{van Mulbregt}, {SciPy 1.0 Contributors}, {SciPy} 1.0: Fundamental algorithms for scientific computing in python, Nature Methods 17 (2020) 261--272.
\newblock \href {https://doi.org/10.1038/s41592-019-0686-2} {\path{doi:10.1038/s41592-019-0686-2}}.

\bibitem{pyade2019}
{xKuZz}, \href{https://pypi.org/project/pyade-python/}{{PyADE}: Python advanced differential evolution algorithms library}, PyPI package, version 1.1, accessed 31 July 2026 (2019).
\newline\urlprefix\url{https://pypi.org/project/pyade-python/}

\bibitem{more2009benchmarking}
J.~J. Mor{\'e}, S.~M. Wild, Benchmarking derivative-free optimization algorithms, SIAM Journal on Optimization 20~(1) (2009) 172--191.
\newblock \href {https://doi.org/10.1137/080724083} {\path{doi:10.1137/080724083}}.

\bibitem{beiranvand2017bestpractices}
V.~Beiranvand, W.~Hare, Y.~Lucet, Best practices for comparing optimization algorithms, Optimization and Engineering 18~(4) (2017) 815--848.
\newblock \href {https://doi.org/10.1007/s11081-017-9366-1} {\path{doi:10.1007/s11081-017-9366-1}}.

\bibitem{hansen2021coco}
N.~Hansen, A.~Auger, R.~Ros, O.~Mersmann, T.~Tu{\v{s}}ar, D.~Brockhoff, {COCO}: A platform for comparing continuous optimizers in a black-box setting, Optimization Methods and Software 36~(1) (2021) 114--144.
\newblock \href {https://doi.org/10.1080/10556788.2020.1808977} {\path{doi:10.1080/10556788.2020.1808977}}.

\bibitem{finck2009bbob}
S.~Finck, N.~Hansen, R.~Ros, A.~Auger, \href{https://inria.hal.science/inria-00362633v2/document}{Real-parameter black-box optimization benchmarking 2009: Noiseless functions definitions}, Tech. Rep. RR-6829, INRIA, updated version as of February 2019 (2009).
\newline\urlprefix\url{https://inria.hal.science/inria-00362633v2/document}

\bibitem{hansen2009bbobsetup}
N.~Hansen, A.~Auger, S.~Finck, R.~Ros, \href{https://inria.hal.science/inria-00362649v2/document}{Real-parameter black-box optimization benchmarking 2009: Experimental setup}, Tech. Rep. RR-6828, INRIA (2009).
\newline\urlprefix\url{https://inria.hal.science/inria-00362649v2/document}

\bibitem{jamil2013benchmarkfunctions}
M.~Jamil, X.-S. Yang, A literature survey of benchmark functions for global optimisation problems, International Journal of Mathematical Modelling and Numerical Optimisation 4~(2) (2013) 150--194.
\newblock \href {https://doi.org/10.1504/IJMMNO.2013.055204} {\path{doi:10.1504/IJMMNO.2013.055204}}.

\bibitem{derrac2011nonparametric}
J.~Derrac, S.~Garc{\'i}a, D.~Molina, F.~Herrera, A practical tutorial on the use of nonparametric statistical tests as a methodology for comparing evolutionary and swarm intelligence algorithms, Swarm and Evolutionary Computation 1~(1) (2011) 3--18.
\newblock \href {https://doi.org/10.1016/j.swevo.2011.02.002} {\path{doi:10.1016/j.swevo.2011.02.002}}.

\bibitem{garcia2010nonparametric}
S.~Garc{\'i}a, A.~Fern{\'a}ndez, J.~Luengo, F.~Herrera, Advanced nonparametric tests for multiple comparisons in the design of experiments in computational intelligence and data mining: Experimental analysis of power, Information Sciences 180~(10) (2010) 2044--2064.
\newblock \href {https://doi.org/10.1016/j.ins.2009.12.010} {\path{doi:10.1016/j.ins.2009.12.010}}.

\bibitem{hansen2022anytime}
N.~Hansen, A.~Auger, D.~Brockhoff, T.~Tu{\v{s}}ar, Anytime performance assessment in blackbox optimization benchmarking, IEEE Transactions on Evolutionary Computation 26~(6) (2022) 1293--1305.
\newblock \href {https://doi.org/10.1109/TEVC.2022.3210897} {\path{doi:10.1109/TEVC.2022.3210897}}.

\bibitem{choi2025eigenvector}
T.~J. Choi, An efficient eigenvector-based crossover for differential evolution: Simplifying with rank-one updates, AIMS Mathematics 10~(2) (2025) 3500--3522.
\newblock \href {https://doi.org/10.3934/math.2025162} {\path{doi:10.3934/math.2025162}}.

\bibitem{illesova2025vhaoptimization}
S.~Ill{\'e}sov{\'a}, V.~Nov{\'a}k, T.~Bezd{\v{e}}k, C.~Possel, M.~Beseda, Numerical optimization strategies for the variational hamiltonian ansatz in noisy quantum environments (2025).
\newblock \href {http://arxiv.org/abs/2505.22398} {\path{arXiv:2505.22398}}, \href {https://doi.org/10.48550/arXiv.2505.22398} {\path{doi:10.48550/arXiv.2505.22398}}.

\bibitem{novak2025reliablevqa}
V.~Nov{\'a}k, S.~Ill{\'e}sov{\'a}, T.~Bezd{\v{e}}k, I.~Zelinka, M.~Beseda, Reliable optimization under noise in quantum variational algorithms (2025).
\newblock \href {http://arxiv.org/abs/2511.08289} {\path{arXiv:2511.08289}}, \href {https://doi.org/10.48550/arXiv.2511.08289} {\path{doi:10.48550/arXiv.2511.08289}}.

\bibitem{novak2026vqaoptimizers}
V.~Nov{\'a}k, I.~Zelinka, V.~Sn{\'a}{\v{s}}el, Optimization strategies for variational quantum algorithms in noisy landscapes, Evolutionary IntelligenceAccepted for publication; preprint available as arXiv:2506.01715 (2026).
\newblock \href {http://arxiv.org/abs/2506.01715} {\path{arXiv:2506.01715}}.

\end{thebibliography}

\end{document}